%% file: example_paper.tex
\documentclass{article}

\usepackage{microtype}
\usepackage{graphicx}
\usepackage{subcaption}
\usepackage{booktabs} 
\usepackage{xcolor} 
\usepackage{algorithm} 
\colorlet{DarkBlue}{blue!70!black}
\colorlet{DarkRed}{red!70!black}
\usepackage[table]{xcolor}
\usepackage[most]{tcolorbox}
\usepackage{paralist}
\usepackage{multirow}

\usepackage{hyperref}
\definecolor{lightblue}{RGB}{200,220,255}
\definecolor{redtext}{RGB}{255,0,0}

\usepackage[preprint]{icml2026}

\usepackage{amsmath}
\usepackage{amssymb}
\usepackage{mathtools}
\usepackage{amsthm}

\usepackage[capitalize,noabbrev]{cleveref}

\theoremstyle{plain}

\theoremstyle{definition}

\theoremstyle{remark}

\usepackage[textsize=tiny]{todonotes}

\begin{document}

\onecolumn
  \icmltitle{Anticipation-VLA: Solving Long-Horizon Embodied Tasks via Anticipation-based Subgoal Generation}



  \icmlsetsymbol{equal}{*}
  \icmlsetsymbol{corres}{$\dagger$}

  \begin{icmlauthorlist}
    \icmlauthor{Zhilong Zhang}{n1,n2,equal}
    \icmlauthor{Wenyu Luo}{n2,equal}
    \icmlauthor{Haonan Wang}{n2,equal}
    \icmlauthor{Yifei Sheng}{n2,equal}
    \icmlauthor{Yidi Wang}{n1,n2}
    \icmlauthor{Hanyuan Guo}{n2}
    \icmlauthor{Haoxiang Ren}{n2}
    \icmlauthor{Xinghao Du}{n1,n2}
    \icmlauthor{Yuhan Che}{n3}
    \icmlauthor{Tongtong Cao}{n3}
    \icmlauthor{Lei Yuan}{n1,n2}
    \icmlauthor{Yang Yu}{n1,n2,corres}

  \end{icmlauthorlist}

  \icmlaffiliation{n1}{National Key Laboratory for Novel Software Technology, Nanjing University, Nanjing, China}
  \icmlaffiliation{n2}{School of Artificial Intelligence, Nanjing University, Nanjing, China}
  \icmlaffiliation{n3}{Department of Foundation model, 2012 Labs, Huawei}

  \icmlcorrespondingauthor{Yang Yu}{yuy@nju.edu.cn}

  \icmlkeywords{Machine Learning, ICML}

  \vskip 0.3in



\printAffiliationsAndNotice{}  

\begin{abstract}
Vision-Language-Action (VLA) models have emerged as a powerful paradigm for embodied intelligence, enabling robots to perform tasks based on natural language instructions and current visual input. However, existing VLA models struggle with long-horizon tasks due to compounding errors. Prior methods decompose tasks into subtasks of fixed granularity, which cannot adapt to the varying complexity of execution states, limiting their robustness in long-horizon tasks. To overcome this, we introduce \textbf{Anticipation Model}, which \textit{adaptively and recursively generates future subgoals}. This model continuously adapts as the task unfolds, adjusting future subgoals in response to evolving dynamics, facilitating more reliable planning paths. Building on this concept, we propose \textbf{Anticipation-VLA}, a hierarchical VLA model that leverages the anticipation model to generate actionable subgoals that guide VLA policy execution. We implement Anticipation-VLA with finetuning a Unified Multimodal Model (UMM) for high-level subgoal generation and a goal-conditioned VLA policy for low-level action execution. Experiments in both simulated and real-world robotic tasks demonstrate the effectiveness of Anticipation-VLA, highlighting the importance of adaptive and recursive subgoal generation for robust policy execution.

\end{abstract}

\section{Introduction}
\begin{figure*}[htbp]
  \centering
  \includegraphics[width=0.92\linewidth]{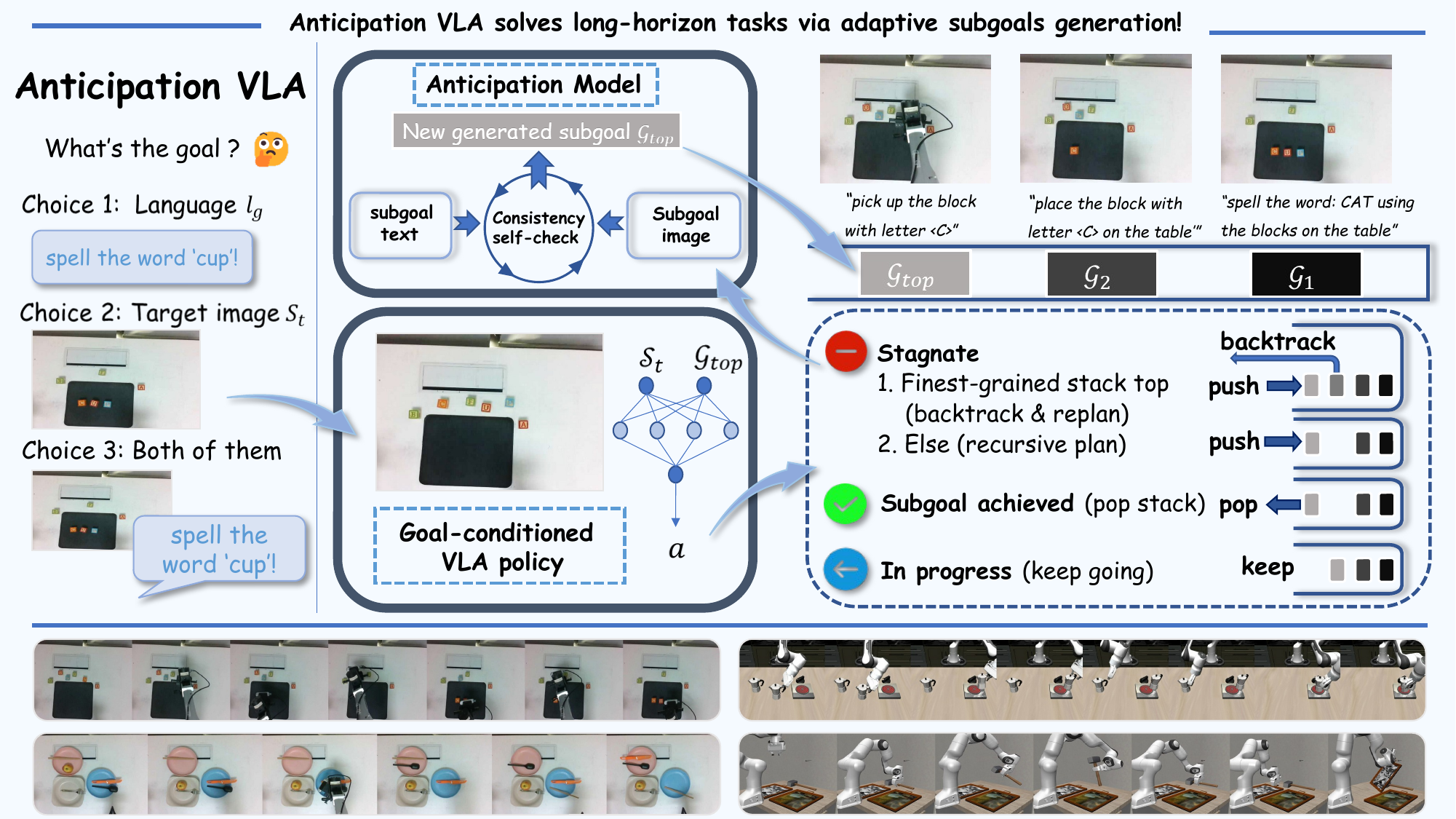}
  \caption{Overall architecture of Anticipation-VLA. The anticipation model adaptively outputs multimodal subgoals guided by progress feedback, while a goal-conditioned VLA executes low-level actions. During this process, we maintain a dynamic goal stack enables backtracking and refinement for robust execution.}
  \label{fig:overview}
  \vspace{-4mm}
\end{figure*}

Vision-Language-Action (VLA) models have emerged as a powerful paradigm for embodied intelligence, allowing robots to execute low-level actions directly from raw visual observations and natural language instructions~\citep{rt12022arxiv,rt22023arxiv,octo_2023,kim2024openvla,black2024pi_0,kim2025fine}. By leveraging large-scale robots data pretraining, these models demonstrate impressive generalization across tasks, objects, and environments~\citep{team2025gemini,intelligence2025pi05visionlanguageactionmodelopenworld,intelligence2025pi}. 

Despite these advancements, existing VLA models face significant challenges when dealing with long-horizon tasks~\citep{fan2025long,zhang2025vlabench,gao2025vlaos,li2025simplevla} . While prior research has explored approaches such as skill retrieval~\citep{long2025checkmanual,fan2025long}, VLM-based planning~\citep{mu2023embodiedgpt,brohan2023can,intelligence2025pi05visionlanguageactionmodelopenworld}, and subgoal image prediction~\citep{blackzero,wu2024robomind,sun2025planning,gao2025vlaos}, most of these methods rely on fixed-granularity subtask planning. As a result, they often generate either overly fine-grained subgoals that introduce unnecessary complexity, or too sparse subgoal generation that fail to guide policy execution, ultimately hindering scalability and generalization.

To address these, we introduce \textbf{Anticipation Model} \cite{yu25} for adaptive and recursive subgoal generation. Given the current observation and the desired goal, the anticipation model outputs the multimodal subgoal that lies on the optimal path from the current state to the goal objective, thus breaking down high-level instructions into intermediate achievable goals. Its core property is \textbf{recursive generation}: each generated subgoal can be further decomposed into more precise subgoals once policy execution is staged. By continuously updating subgoals based on current observations and task progress, the model provides a principled approach for long-horizon task decomposition. This progress-aware decomposition ensures that each subgoal contributes meaningfully to maximizing cumulative reward, while simplifying policy execution across arbitrary horizons.

Building on this conceptual model, we present a hierarchical VLA framework, \textbf{Anticipation-VLA}, that leverages the anticipation model to guide low-level goal-conditioned policy execution, as shown in Figure~\ref{fig:overview}. In this framework, a Unified Multimodal Model (UMM)~\citep{deng2025emerging,sun2025planning,cui2025emu3} serves as both anticipation and value models, reasoning over visual and linguistic inputs to generate subgoals and provide the value estimation. A separate goal-conditioned VLA model executes low-level actions conditioned on these generated subgoals. By monitoring task progress and refining subgoals when progress stalls, Anticipation-VLA enables robust and adaptive execution of long-horizon tasks in complex environments. Experiments in both simulated and real-world robotic tasks demonstrate the effectiveness of Anticipation-VLA, highlighting the importance of dynamic anticipation and recursive subgoal generation for robust policy execution. 

In summary, our contributions are:
\begin{itemize}
\item We introduce the Anticipation Model, an adaptive planning model for value-optimal and policy-aware subgoal generation.
\item We propose Anticipation-VLA, a hierarchical framework that integrates the anticipation model with a goal-conditioned VLA policy for long-horizon task planning and execution.
\item We validate the framework in both simulated and real-world robotic tasks, demonstrating the effectiveness of our hierarchical framework, Anticipation-VLA.
\end{itemize}

\section{The Foundation of Anticipation-VLA}
In this section, we present the foundational framework for the Anticipation-VLA. We begin by formalizing the task environment using the Goal-Conditioned Markov Decision Process (GMDP) in Section~\ref{sec:hg-mdp}. We then introduce the crucial high-level planner Anticipation Model in Section~\ref{sec:concept_anticipation}, followed by the complete architecture and inference procedure for the Anticipation-VLA in Section~\ref{sec:concet_Anticipation-VLA}.

\subsection{Goal-Conditioned Markov Decision Process}
\label{sec:hg-mdp}
While VLA models offer a powerful paradigm for human-robot interaction, scaling to long-horizon tasks presents a major challenge due to exacerbated policy compounding error and limited generalization ability. To address this challenge, we first formalize the problem using the Goal-Conditioned Markov Decision Process (GMDP)~\citep{mdp,puterman2014markov}, which is defined by the tuple $(\mathcal{S}, \mathcal{A}, \mathcal{G}, P, r, T)$.

The state space $\mathcal{S}$ represents the set of all possible visual observations (e.g., camera images), and the action space $\mathcal{A}$ is the set of continuous control inputs the robot can execute. Crucially, the goal space $\mathcal{G}$ is defined as the Cartesian product of the state space and the language space: 
\begin{equation}
\mathcal{G} = (\mathcal{L} \cup \{\emptyset_\mathcal{L}\}) \times (\mathcal{S} \cup \{\emptyset_\mathcal{S}\})  \setminus \{(\emptyset_\mathcal{L}, \emptyset_\mathcal{S})\},
\end{equation}
where $\mathcal{L}$ represents the set of possible instructions, $\emptyset_\mathcal{L}$ is the null instruction, and $\emptyset_\mathcal{S}$ is the null state observation. The exclusion of the non-informative pair $(\emptyset_\mathcal{L}, \emptyset_\mathcal{S})$ ensures that every goal $g \in \mathcal{G}$ is non-vacuous. Under this definition, a goal $g$ is an observation-instruction pair $(\ell_g, s_g)$, where at least one component, $s_g \in \mathcal{S}$ (the target visual state) or $\ell_g \in \mathcal{L}$ (the text command), is explicitly defined. For example, a goal can be purely language-conditioned, such as $(\emptyset_\mathcal{S}$, "Make a coffee"), or purely observation-conditioned, such as ("Image of a cup of coffee", $\emptyset_\mathcal{L}$).

The dynamics function $P: \mathcal{S} \times \mathcal{A} \to \Delta(\mathcal{S})$ defines the probability of transitioning to a new state $s'$ from a current state $s$ after taking an action $a$. The reward function $r: \mathcal{S} \times \mathcal{A} \times \mathcal{G} \to \mathbb{R}$ provides a scalar reward signal for executing action $a$ in state $s$ while pursuing goal $g$. Here, we define the reward as the improvement to the goal from state $s$ after taking action $a$. The objective in a GMDP is to find an optimal goal-conditioned policy $\pi: \mathcal{S} \times \mathcal{G} \to \Delta(\mathcal{A})$ that maximizes the expected cumulative reward $V(\pi)= \mathbb{E}_\pi \left[ \sum_{t=0}^{T} r(s_t, a_t, g)\right]$. Intuitively, this value function can be viewed as the expected total distance the policy needs to traverse from the current observation to the goal. This framework explicitly incorporates the goal into the decision-making process, allowing the policy to adapt its behavior based on the specific task objective defined by the combined visual and linguistic components.

\subsection{Anticipation Model}
\label{sec:concept_anticipation}

\textbf{Formal Definition.} The anticipation model $G$, first introduced by \citet{yu25}, is the central mechanism for enabling hierarchical planning in the GMDP. Its role is to select the immediate subgoal $g' \in \mathcal{G}$ that serves as an efficient stepping stone toward achieving the overall high-level goal $g$. This process directly mitigates compounding error of the subsequent goal-conditioned policy by reducing the effective horizon.

Formally, the anticipation model is a mapping $G: \mathcal{S} \times \mathcal{G} \rightarrow \mathcal{G}$ that takes the current state $s$ and the active goal $g$ as input, and outputs a refined subgoal $g' \in \mathcal{G}$. Here, the recursive nature of the anticipation model arises because the output subgoal $g'$ can be immediately fed back into the model as the new active goal for the next decomposition step, enabling subgoal generation at various granularities. This iterative process continues through multiple intermediate steps until an executable target is obtained.

\textbf{Optimal Subgoal Generation.} The selection of an optimal subgoal $g'$ is governed by the \textit{global optimality of reinforcement learning with anticipation}~\citep{yu25}. This principle requires the \textit{shortest-path reward structure}~\citep{yu25}. The definition of optimality relies on the \textit{Optimal Value Function} $V^{\ast}(s, g)$, which represents the maximum expected cumulative reward from state $s$ towards goal $g$. $V^{\ast}(s, g)$ is the unique solution to the Bellman Optimality Equation:
\begin{equation}
V^{\ast}(s, g) = \max_{a} \left[ r(s,a,g) + \mathbb{E}_{s' \sim P(\cdot|s,a)}[V^{\ast}(s', g)] \right]
\end{equation}

Based on this framework, the anticipation model $G$ is trained to generate a subgoal $g'$ that approaches the following \textit{Optimal Decomposition}:
\begin{equation}
V^{\ast}(s_0, g) = V^{\ast}(s_0, g') + V^{\ast}(s_{g'}, g),
\end{equation}
This equation establishes that the maximum reward from $s_0$ to the final goal $g$ must be perfectly decomposable into the maximum reward from $s_0$ to the waypoint $g'$ plus the maximum reward from the waypoint $g'$ to the final goal $g$. This provides the principled objective for training an optimal intermediate subgoal generator.

\subsection{Anticipation-VLA}
\label{sec:concet_Anticipation-VLA}
The Anticipation-VLA integrates high-level planning and low-level control into a unified, hierarchical system tailored for long-horizon embodied tasks. It consists of three core components: 
\begin{inparaenum}
    \item [1)] \textbf{Anticipation Model ($G$)}: a high-level planner that recursively generates intermediate subgoals.
    \item [2)] \textbf{Optimal Value Function ($V^\ast$)}: a value function that captures the optimal expected cumulative reward to guide subgoal generation.
    \item [3)] \textbf{Goal-Conditioned VLA Model ($\pi$)}: a low-level controller that executes fine-grained actions based on the current observation and the immediate subgoal.
\end{inparaenum}

\textbf{Dynamic Subgoal Management.} The key innovation of the Anticipation-VLA lies in its  adaptive subgoal management. Rather than relying on fixed planning intervals or handcrafted decomposition, the system uses an estimate of the optimal value function $V^\ast$ to dynamically trigger re-planning when progress stalls or a subgoal is achieved. This ensures computational efficiency and robustness in complex environments. The complete inference procedure is detailed in \textbf{Algorithm}~\ref{alg:anticipation_vla_inference}. Specifically, the system maintains a subgoal stack and, at each planning check (every $K$ timesteps), the system evaluates three conditions using $V^\ast$:
\begin{enumerate}
    \item \textbf{Goal Achievement}: if the current value $V^\ast(s, g)$ is close to the value at the goal state ($|V^\ast(s,g)-V^\ast(g,g)|< \delta$), the subgoal is considered complete and popped from the stack.
    \item \textbf{Insufficient Progress}: if the value improvement over the last interval is negligible ($|V^\ast(s,g)-V^\ast(s_\text{prev},g)|< \delta$) and the stack is not full, the anticipation model $G$ generates a refined subgoal and push it into the stack. Moreover, if the stack has reached its maximum depth, the system backtracks to the initial state, as the policy may be stuck in a local stagnation.
    \item \textbf{Sufficient Progress}: otherwise, if substantial progress has been made ($|V^\ast(s,g)-V^\ast(s_\text{prev},g)|\geq \delta$)), the current subgoal remains on the stack, and the process continues until the next evaluation step.
\end{enumerate}
This closed-loop interaction between $G$, $\pi$, and $V^\ast$ enables flexible, error-resilient execution of long-horizon tasks.

\section{Practical Implementation}

\subsection{UMM-based Anticipation Model and Value Model}
We implement our anticipation model based on Bagel~\citep{deng2025emerging}, a UMM model capable of jointly understanding and generating both textual and visual content. Therefore, this model can function simultaneously as an anticipation model and an optimal value model within a unified framework. Simplifying the system design, and enabling the cross-modal knowledge transfer within the shared model.

\textbf{Anticipation Model.} Although UMM can predict a subgoal $g'$ directly from the current observation $s$ and high-level goal $g$, we find that such direct generation often suffers from hallucination, producing subgoals that are inconsistent with the environment state or task semantics. To address this, we decompose the anticipation process into two stages inspired by~\citet{sun2025planning}. First, UMM functions as a language-based policy $ l_\theta:\mathcal{S}\times\mathcal{G}\rightarrow\mathcal{L}$ that predicts a subgoal instruction $\ell_{g'}$ given $(s, g)$. Then, conditioned on $(s, \ell_{g'})$, UMM serves as a dynamics model $P_\theta:\mathcal{S}\times\mathcal{L}\rightarrow\mathcal{S}$ that grounds the instruction and predicts the corresponding subgoal image $s_{g'}$. Formally, we define the anticipation mapping as $G_\theta=(P_\theta\circ l_\theta, l_\theta)$. This decomposition constrains the generation process through a semantically meaningful bottleneck for subgoal generation.

\textbf{Self-Discriminative Regularization.} To further reduce the hallucination, we adopt the self-discriminative regularization introduced in Uni-Plan~\cite{sun2025planning}, which encourages the model to assess the plausibility of its own outputs. Specifically, after generating a candidate subgoal image $s_{g'}$ from $(s, l_{g'})$ by $P_\theta$, we apply the inverse dynamics model of the UMM, denoted as $P_\theta^{-1}: \mathcal{S} \times \mathcal{S} \to \mathcal{L}$, to infer the instruction $l'_\text{inv}$ that would lead from $s$ to $s_{g'}$. If the inferred instruction $l'_\text{inv}$ is semantically equivalent to the original instruction $l_{g'}$, we retain the generated subgoal. Otherwise, we discard it and prompt the anticipation model to regenerate a new candidate.

\textbf{Optimal Value Model.} The standard formulation of an optimal value model takes the current observation and goal as input and predicts the maximum expected cumulative reward. This model is typically trained by Temporal Difference (TD) methods~\citep{sutton}. However, in real-world settings, obtaining step-wise dense rewards is nearly infeasible. Instead, we usually only have the sparse, trajectory-level signals indicating whether a task was successful. Such sparsity destabilizes TD learning and makes it difficult to train a reliable and accurate value model.

Fortunately, in our system, the optimal value model is not tasked with computing the exact absolute value but rather with determining \textit{Goal Achievement} or \textit{Progress Stagnation} as shown in Algorithm~\ref{alg:anticipation_vla_inference}. This allows us to reformulate the value prediction task as a \textbf{classification problem}. Specifically, the value model $V_\theta$ takes the current observation $s$, the previous observation $s_\text{prev}$, and the goal $g$ as input, and classifies the state as one of the following three categories: goal achieved, progress stagnated, or progress improved. This transformation simplifies the original value regression task into a more manageable classification task. More details are shown in Appendix~\ref{app:ant_data}.

\begin{figure}[htbp]
    \centering
    \includegraphics[width=\linewidth]{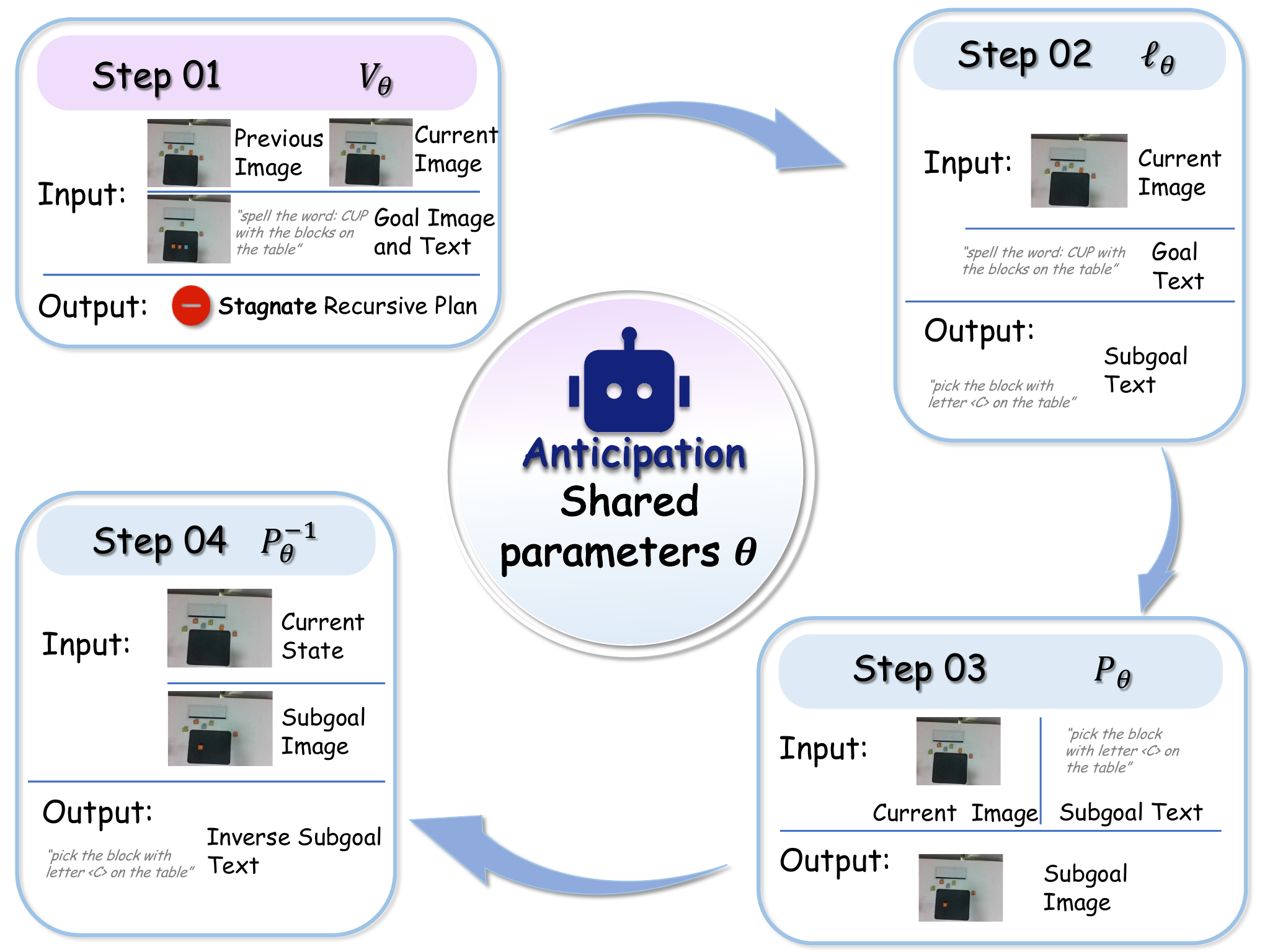}
    \caption{Inference procedure of Anticipation Model. At each inference step, \textit{Value Model} $V_\theta$ first outputs a progress label. If stagnation is detected, \textit{Policy Model} $\pi_\theta$ generates a textual subgoal. \textit{Dynamics Model} $P_\theta$ then predicts the corresponding visual subgoal. Finally, \textit{Inverse Dynamics Model} $P_\theta^{-1}$ verifies whether the actions inferred from the predicted visual transition align with the generated textual subgoal.}
    \label{fig:inference}
    \vspace{-1mm}
\end{figure}

\textbf{Data Preparation.}
We construct two hierarchical datasets to train the anticipation model and the value model, respectively:  
(i)~$\mathcal{D}_\text{anti} = \bigcup_{h=1}^H \mathcal{D}_\text{anti}^h$, where each $\mathcal{D}_\text{anti}^h = \{(s, g_{h+1} = (\ell_{g_{h+1}}, s_{g_{h+1}}), g_h)\}$ consists of tuples of a current observation $s$, a current goal $g_h$, and a sub-level goal $g_{h+1}$.
(ii)~$\mathcal{D}_\text{value} = \bigcup_{h=1}^H \mathcal{D}_\text{value}^h$, where each $\mathcal{D}_\text{value}^h = \{(s_1, s_2, g_h, y)\}$ contains pairs of sequential observations $(s_1, s_2)$, the goal $g_h$, and the progress label $y \in \{\text{progress}, \text{achieve}, \text{no progress}\}$ assigned based on temporal and geometric proximity to the goal. Here, $H$ denotes the total number of subgoal hierarchy levels, with level $h=1$ corresponding to the highest-level goals (original task description) and level $h=H$ to the lowest-level (finest-grained) subgoals. Full details of the labeling protocol and sampling strategy are shown in the Appendix~\ref{app:data_construction}.

\textbf{Training Recipe.} In general, our anticipation model requires optimizing four distinct roles: (i) the language-based policy $ l_\theta$, (ii) the forward dynamics model $P_\theta$, (iii) the inverse dynamics model $P_\theta^{-1}$, and (iv) the optimal value model $V_\theta$. The corresponding loss functions optimized over the dataset $\mathcal{D}_\text{anti}$ and $\mathcal{D}_\text{value}$ are defined below:
\begin{itemize}
    \item \textbf{Policy Loss}: A cross-entropy (CE) loss for training the policy model to predict the subgoal instruction $\ell_{g_{h+1}}$.
    \begin{equation}
    \mathcal{L}_\text{policy}(\theta) = \mathbb{E}_{\mathcal{D}_\text{anti}}[-\log  l_\theta(\ell_{g_{h+1}} | s, g_h)],
    \end{equation}
    
    \item \textbf{Forward Dynamics Loss}: A mean-squared-error (MSE) loss for training the flow matching model to predict the next state $s_{g_{h+1}}$. This minimizes the error between the predicted velocity $v_\theta$ and the actual sampled velocity $v$.
    \begin{equation}
    \mathcal{L}_\text{dyna}(\theta) = \mathbb{E}_{t, \mathcal{D}_\text{anti}}[||v_\theta(s, s^t_{g_{h+1}}, \ell_{g_{h+1}}, t) - v||_2^2],
    \label{loss:dynamics}
    \end{equation}

    \item \textbf{Inverse Dynamics Loss}: A CE loss for training the inverse dynamics model to predict the instruction $l_{g_{h+1}}$ that connects $s$ and $s_{g_{h+1}}$.
    \begin{equation}
    \mathcal{L}_\text{inverse}(\theta) = \mathbb{E}_{\mathcal{D}_\text{anti}}[-\log P_{\theta}^{-1}(\ell_{g_{h+1}}| s, s_{g_h})],
    \label{loss:inverse}
    \end{equation}

    \item \textbf{Value Loss}: A CE loss that trains the value model to predict goal achievement or progress status.
    \begin{equation}
    \mathcal{L}_\text{value}(\theta) = \mathbb{E}_{\mathcal{D}_\text{value}}[-\log V_{\theta}(y| s_1, s_2, g_h)],
    \label{loss:value}
    \end{equation}
\end{itemize}

Notably, all of these models are implemented within the unified UMM architecture, where the overall loss function is defined as a weighted sum of the loss components:
\begin{equation}
\begin{split}
\mathcal{L}(\theta)=&\lambda_1 \mathcal{L}_\text{policy}(\theta)+\lambda_2  \mathcal{L}_\text{dyna}(\theta) \\
+ &\lambda_3  \mathcal{L}_\text{inverse}(\theta)+\lambda_4  \mathcal{L}_\text{value}(\theta),
\end{split}
\end{equation}

\subsection{Goal-conditioned VLA Model}
We implement the goal-conditioned VLA model based on \(\pi_{0.5}\)~\citep{intelligence2025pi05visionlanguageactionmodelopenworld}, a state-of-the-art flow matching-based VLA model. Specifically, we enhance the VLA model by incorporating a current subgoal image \( s_g^t \) as part of the input, placing it after the current observation from all cameras, \(\mathbf{s}_o^t = [s_1^t, \dots, s_n^t]\). Subsequently, we concatenate the robot’s configuration \(\mathbf{q}\) and the subgoal instruction \(\ell_g^t\) after the images. The goal-conditioned VLA model captures the distribution \(\pi_\theta(\mathbf{a}^{t:t+h}|\mathbf{s}_o^t, g_t = (s_g, \ell_g))\), where \( g_t \) represents the subgoal at time \( t \).

Furthermore, to address potential issues such as ambiguity and deformation in the generated subgoals from the anticipation model, we improve the model's robustness to goal image variations by randomly masking out the tokens of the goal image \( s_g \) during training. This regularization technique ensures better generalization in the presence of noisy or incomplete goal representations.

Given an overall goal \( g \), which can either be a language instruction or a goal image, we combine the UMM-based anticipation model and the goal-conditioned VLA model. Our Anticipation-VLA framework decomposes the distribution as follows:
\begin{equation}
    \pi_\theta(\mathbf{a}^{t:t+h}|\mathbf{s}_o^t, g) = \pi_\theta(\mathbf{a}^{t:t+h}|\mathbf{s}_o^t, g_t) \cdot G_\theta(g_t | g)
\end{equation}
where \( \pi_\theta(\mathbf{a}^{t:t+h}|\mathbf{s}_o^t, g_t) \) is the action distribution conditioned on the observations and subgoal, and \( G_\theta(g_t | g) \) represents the anticipation model output probability of the goal \( g_t \) given the overall goal \( g \).

\textbf{Data Preparation.} We construct a hierarchical dataset $\mathcal{D}_\text{policy} = \bigcup_{h=1}^H \mathcal{D}_\text{policy}^h$ for training the goal-conditioned VLA model, where each level $h$ contains tuples $\mathcal{D}_\text{policy}^h = \{(s, a, g_h)\}$. Here, $s$ denotes the current observation, $a$ is an action chunk, and $g_h = (\ell_{g_h}, s_{g_h})$ represents the corresponding subgoal at hierarchy level $h$. We reuse the same annotated subgoals from the anticipation dataset to ensure consistency between policy training and subgoal prediction.

\textbf{Training Recipe.} Our goal-conditioned VLA model uniformly samples data from $\mathcal{D}_\text{policy}$ and is finetuned using a flow matching loss. To improve robustness against inaccurate or noisy subgoals produced by the anticipation model at inference time, we randomly mask the goal inputs with a fixed probability during VLA training.

\section{Experiments}
In this section, we perform comprehensive experiments to answer the following questions:
\begin{inparaenum}
    \item[\textbf{Q1}:] How does Anticipation-VLA perform in simulated tasks? (Section~\ref{sec:exp_sim})
    \item[\textbf{Q2}:] How does Anticipation-VLA perform in real-world tasks? (Section~\ref{sec:exp_real})
    \item[\textbf{Q3}:] How sensitive is Anticipation-VLA to hyperparameters and components design choices? (Section~\ref{sec:exp_ablation})
    \item[\textbf{Q4}:] Can Anticipation-VLA generalize to more challenging unseen tasks? (Section~\ref{sec:exp_unseen})
    \item[\textbf{Q5}:] How well do anticipation models generate subgoal text and images? (Section~\ref{sec:exp_ant})
\end{inparaenum}

\textbf{Benchmarks.}
We evaluate Anticipation-VLA on two widely used robotic manipulation simulation benchmarks. \textbf{Libero}~\citep{liu2023Libero} is a lifelong learning benchmark designed for language-guided manipulation tasks. We test our model and baselines on four task suites, \textit{Goal}, \textit{Spatial}, \textit{Object}, and \textit{Long}, with each suite consisting of 10 tasks. All models are trained in a one-shot setting using 40 trajectories in total. \textbf{VLABench}~\citep{zhang2025vlabench} is a large-scale VLA benchmark focused on long-horizon reasoning and execution. We conduct evaluation on the challenging \textit{Hammer Nail \& Hang Picture} task, where nearly all models fail to achieve any success. This task not only requires precise control but also demands strong reasoning and planning abilities to select the correct picture, which is unseen during training. For this task, we train all models on 100 trajectories.

\subsection{Simulated Experiments}
\label{sec:exp_sim}
\begin{figure}[htbp]
  \centering
  \includegraphics[width=0.98\linewidth]{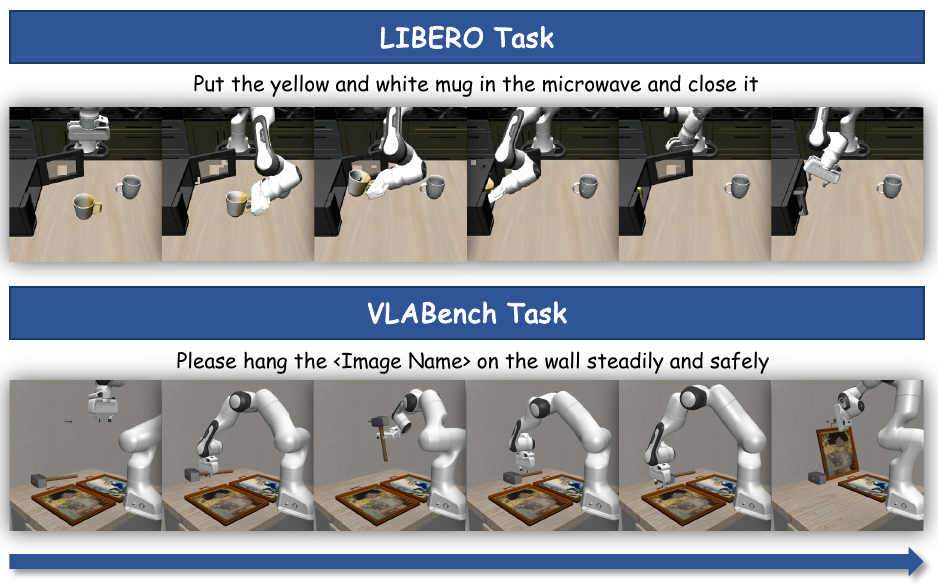}
  \caption{Illustration of Simulated Tasks.}
  \label{fig:}
  \vspace{-4mm}
\end{figure}

\textbf{Baselines.}
We compare Anticipation-VLA against several baselines including:
\begin{inparaenum}
\item[(i)] $\pi_0$~\citep{black2024pi_0}, a VLA model pretrained on large-scale real-world robotic dataset;
\item[(ii)] UniVLA~\citep{wang2025unified}, a unified and native multimodal VLA model that with explicit future image generation; 
\item[(iii)] DreamVLA~\citep{zhangdreamvla}~\footnote{For DreamVLA, we follow the original setup by pretraining on Libero90 and then performing one-shot SFT on Libero.}, a VLA model that integrates comprehensive world knowledge with implicit subgoal prediction;
\item[(iv)] $\pi_{0.5}$~\citep{intelligence2025pi05visionlanguageactionmodelopenworld}, the state-of-art VLA model with subtask prediction pretraining;
\item[(v)] $\pi_{0.5}$+VLM~\footnote{We finetune Qwen2.5-7B to serve as the VLM for subgoal text  generation.}, a variant of $\pi_{0.5}$ enhanced subtask planning ability through VLM model.
\end{inparaenum} We implement all baselines using their official codebases.

\begin{table}[htbp]
\vspace{-1mm}
\centering
\caption{Performance Comparison on Libero.}
\begin{tabular}{lccccc}
\toprule
\multirow{2}{*}{\textbf{Model}} & \multicolumn{5}{c}{\textbf{Libero}} \\
\cmidrule(lr){2-6}
& \textbf{Spatial} & \textbf{Object} & \textbf{Goal} & \textbf{Long} & \textbf{Avg} \\
\midrule
\multicolumn{6}{c}{One-Trajectory SFT} \\
\midrule
$\pi_0$  & 70.2&80.0& 70.6 & 37.6&64.6 \\
UniVLA   & 26.0 & 40.0 & 18.0 & 1.8 & 21.5\\
DreamVLA &38.0 & 34.0 & 16.6 & 20.6 &27.3  \\
\midrule
$\pi_{0.5}$      & 78.2 & 88.6 &  85.8 & 54.6 & 76.8\\
$\pi_{0.5}$+VLM  &  \textbf{82.0} & 88.0 & 80.8 & 53.2 & 76.0 \\
\textbf{Anticipation-VLA}   & 81.8 & \textbf{91.6} & \textbf{86.6} & \textbf{63.2} & \textbf{80.8} \\
\bottomrule
\end{tabular}
\label{tab:Libero_results}
\vspace{-2mm}
\end{table}

\begin{table}[htbp]
\centering
\caption{Performance Comparison on VLABench.}
\begin{tabular}{lcc}
\toprule
\multirow{2}{*}{\textbf{Model}} & \multicolumn{2}{c}{\textbf{VLABench}} \\
\cmidrule(lr){2-3}
& \textbf{Process Reward} & \textbf{Success Rate} \\
\midrule
$\pi_0$ & 39.6 & 1.0 \\
UniVLA & 28.1 &  1.0 \\
DreamVLA & 7.3 & 0.0   \\
\midrule
$\pi_{0.5}$      & 42.7 & 2.1 \\
$\pi_{0.5}$+VLM      & 47.9 & 2.1 \\
\textbf{Anticipation-VLA}   &\textbf{ 56.3 }&\textbf{ 4.2}  \\
\bottomrule
\end{tabular}
\label{tab:vlabench_results}
\end{table}

\textbf{Task Results.}
On Libero (Table~\ref{tab:Libero_results}), Anticipation-VLA demonstrates superior performance across all task suites than either implicit and explicit planning-based baselines. In particular, it achieves the highest improvement on the most challenging long-horizon task \textit{Libero-Long}, demonstrating it effectiveness on managing long-horizon and complex tasks. On VLABench (Table~\ref{tab:vlabench_results}), Anticipation-VLA again surpasses all baselines in both process reward and success rate. The higher process reward suggests that Anticipation-VLA adheres to more coherent, task-relevant intermediate steps throughout execution. Remarkably, it achieves at least twice the success rate of the baseline on the highly demanding \textit{Hammer Nail \& Hang Picture} task. This substantial improvement underscores the effectiveness of our anticipation mechanism in facilitating robust long-horizon reasoning and precise execution even in OOD scenarios.

\subsection{Real-World Experiments}
\label{sec:exp_real}
\begin{figure*}[htbp]
  \centering
  \includegraphics[width=0.98\linewidth]{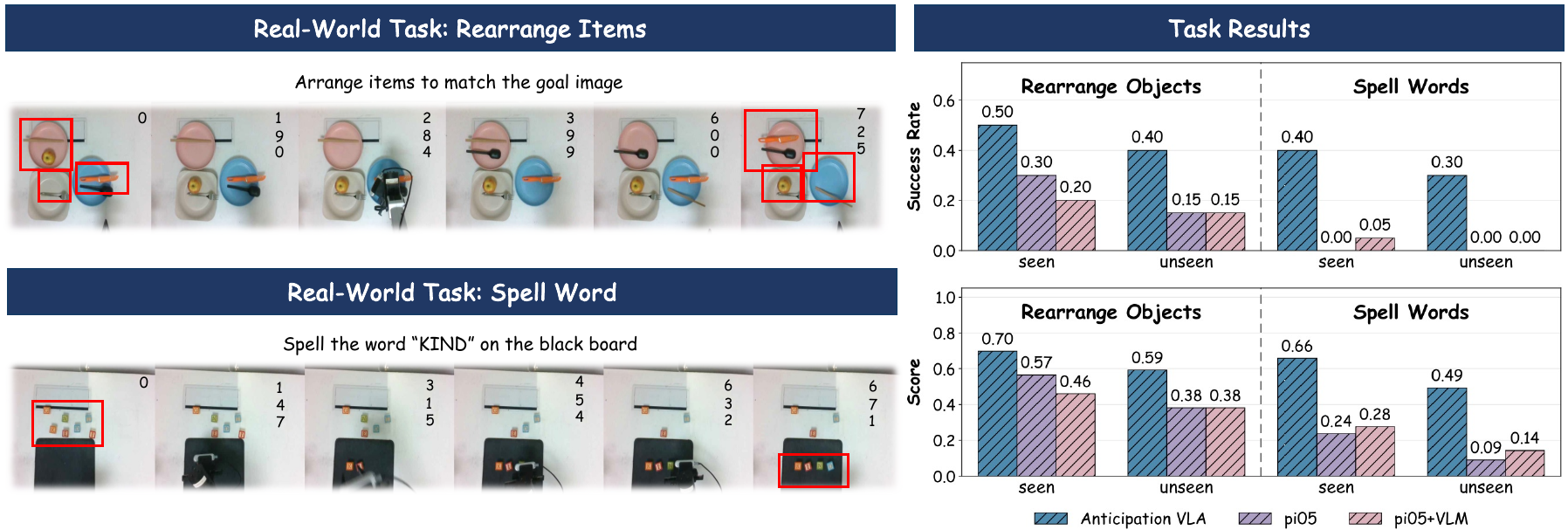}
  \caption{Task illustration and Comparison Results of real-world experiments. }
  \label{fig:real_world}
  \vspace{-5mm}
\end{figure*}

\textbf{Hardware \& Task Design.}
 To evaluate the real-world performance of Anticipation-VLA, we conduct real-world experiments on the Arx-X5 mobile manipulator platform. We design two long-horizon tasks each with a different modality of goal specification, as shown in Figure~\ref{fig:real_world}.
\begin{inparaenum}
\item[1)] \textbf{Rearrange Objects}: VLAs should rearrange physical objects on a table to match a provided goal image.
\item[2)] \textbf{Spell Words}: VLAs should place letter tiles in the correct sequence according to the overall instruction.
\end{inparaenum}
We collect 100 multi-stage expert demonstrations via human teleoperation for \textit{Rearrange Objects} and 200 multi-stage demonstrations for \textit{Spell Words}.

\textbf{Baselines.} We include $\pi_{0.5}$ and $\pi_{0.5}$+VLM, both of which are pretrained on extensive real-robot data and previously shown to outperform standard VLA models in simulated tasks. For evaluation, each model is tested over 40 rollouts per task: 20 on seen and 20 on unseen configurations. We report the trajectory-level \textit{success rate} and the stage-level \textit{score}. Additional details on the hardware setup, task design, and data collection can be found in Appendix~\ref{app:hardware}, Appendix~\ref{app:task}, and Appendix~\ref{app:data}, respectively.

\textbf{Task Results.}
As shown in the right panel of Figure~\ref{fig:real_world}, Anticipation-VLA consistently outperforms all baselines across both real-world tasks, demonstrating improved performance under both image-based and language-based goals. The improvement is especially pronounced in unseen scenarios, where the model exhibits substantially larger gains (\textbf{+107\%}) compared to seen configurations (\textbf{+60\%}). Notably, Anticipation-VLA is the only model achieves non-zero success rate in \textit{unseen Spell Words} task, highlighting its strong generalizability. In contrast, $\pi_{0.5}$ baselines show limited performance even when augmented with a VLM for planning, underscoring that effective long-horizon reasoning relies not on static external modules but on an adaptive multimodal anticipation mechanism. To further illustrate this planning process, we visualize the evolution of the goal stack during inference in Appendix~\ref{app:goal_stack_vis}, revealing how it progressively decomposes goals into actionable steps.


\subsection{Ablation Study}
\label{sec:exp_ablation}

To isolate the contribution of each component in Anticipation VLA, we conduct an ablation study with three variants: (1) \emph{w/o subgoal image}, which removes predicted future frames from the policy input; (2) \emph{w/o subgoal text}, which excludes generated textual subgoals; and
(3) \emph{w/o recursive}, which replaces the adaptive recursive planning with a fixed-level generation strategy.

\begin{figure}[htbp]
  \centering
  \includegraphics[width=1.00\linewidth]{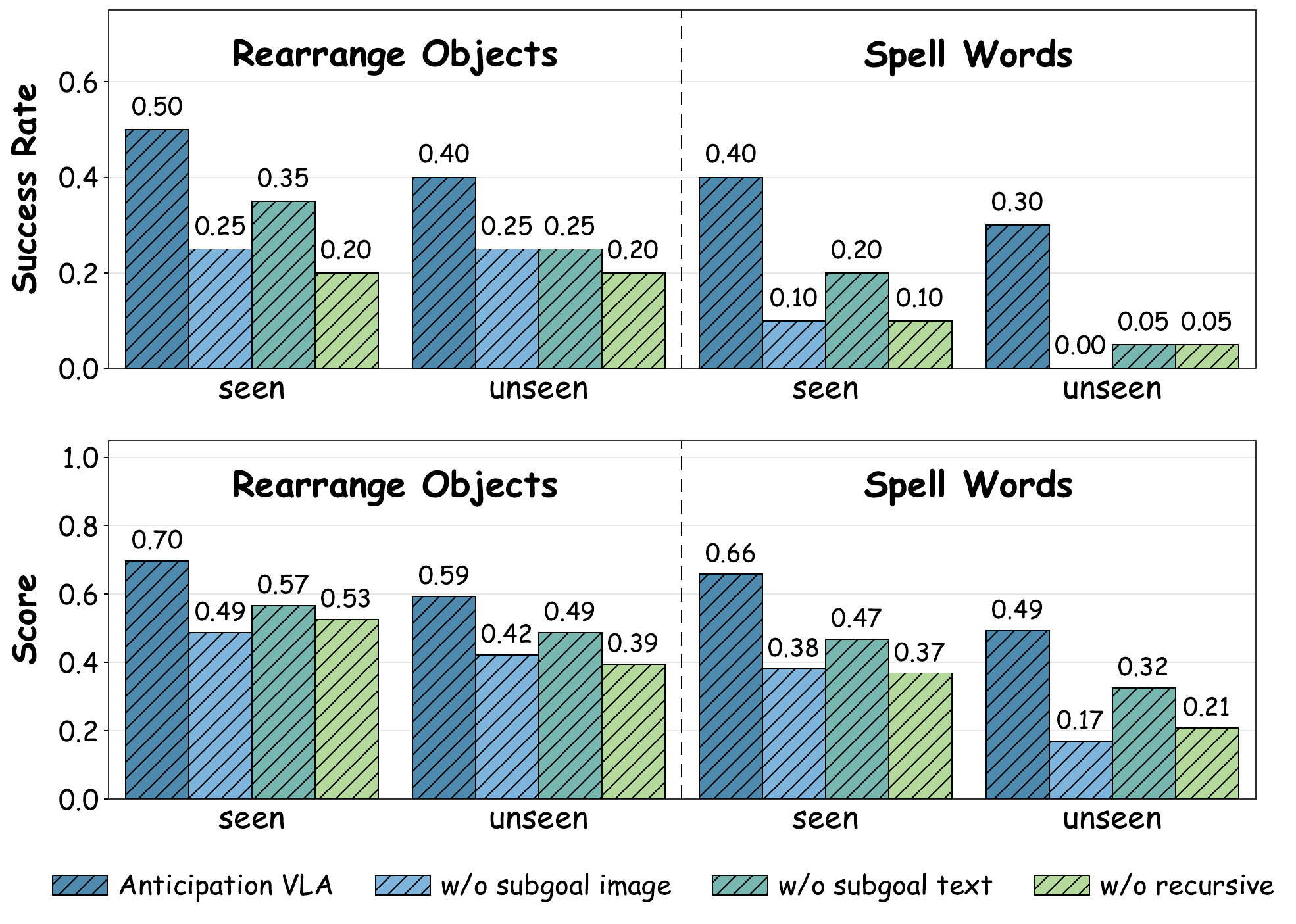}
  \caption{Ablation Study on Real-World Tasks.}
  \label{real-world_ablation}
  \vspace{-3mm}
\end{figure}

\textbf{Task Results.} Figure~\ref{real-world_ablation} summarizes the results across both \textit{Rearrange Objects} and \textit{Spell Words} tasks. As shown in the results, the standard Anticipation-VLA consistently outperforms all ablated variants. This validates the complementarity of our design: adaptive planning supports long-horizon execution, visual subgoals provide physical guidance, while textual predictions offer semantic grounding. Together, the integration of multimodal anticipation and adaptive planning proves essential for robust real-world manipulation.

\subsection{Generalization Evaluation}
\label{sec:exp_unseen}
To rigorously assess the generalization of Anticipation-VLA, we extend our evaluation to two more challenging settings.
\begin{inparaenum}
    \item[(1)] \textbf{Object Generalization}: Testing the model's ability to manipulate novel instances. For instance, in the \textit{Spell Words} task, where the training corpus consists exclusively of alphabetic tiles, we mandate the agent to compose alphanumeric sequences (e.g., ``H2O'').
    \item[(2)] \textbf{Background Generalization}: Evaluating the model's invariance to environmental perturbations. We dramatically alter the visual scene by modifying surface textures and varying illumination conditions. 
\end{inparaenum}
For further details, please refer to Appendix~\ref{app:eval}.

\begin{figure}[htbp]
  \centering
  \includegraphics[width=0.98\linewidth]{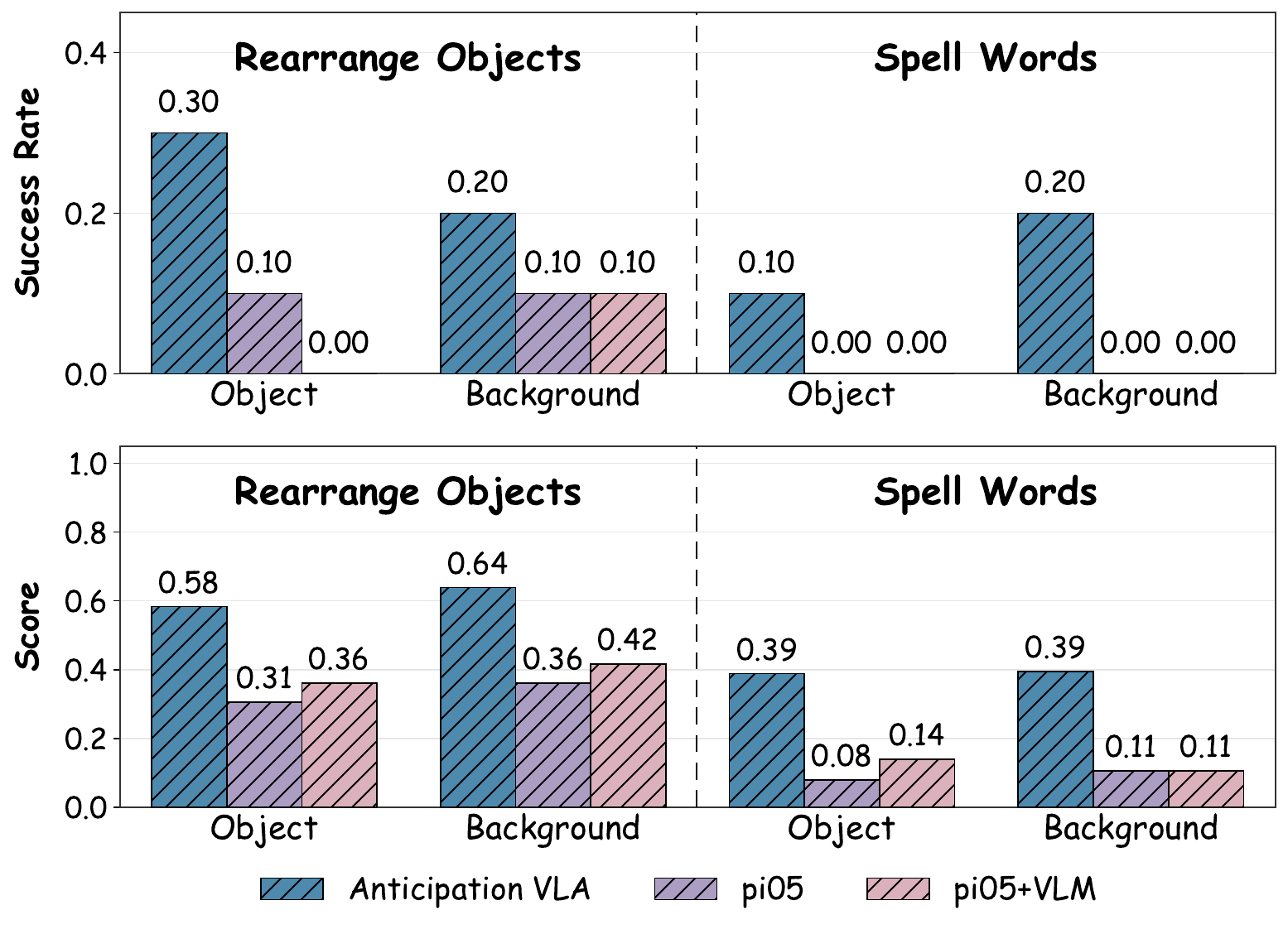}
  \caption{Performance Comparison on Generalization Settings.}
  \label{real-world_generalization}
  \vspace{-2mm}
\end{figure}

\textbf{Task Results.}
Figure~\ref{real-world_generalization} empirically validates the superior robustness of Anticipation-VLA. 
In \textit{Rearrange Objects}, the model exhibits negligible degradation, achieving scores ($0.58$ for objects, $0.64$ for backgrounds) comparable to the standard unseen benchmark ($0.59$) reported in Figure~\ref{fig:real_world}. 
Crucially, in the \textit{Spell Words} task, Anticipation-VLA stands as the only policy to achieve non-zero success, in stark contrast to the complete failure observed in the baselines.

\subsection{Anticipation Evaluation}
\label{sec:exp_ant}
\begin{figure}[htbp]
  \centering
  \includegraphics[width=\linewidth]{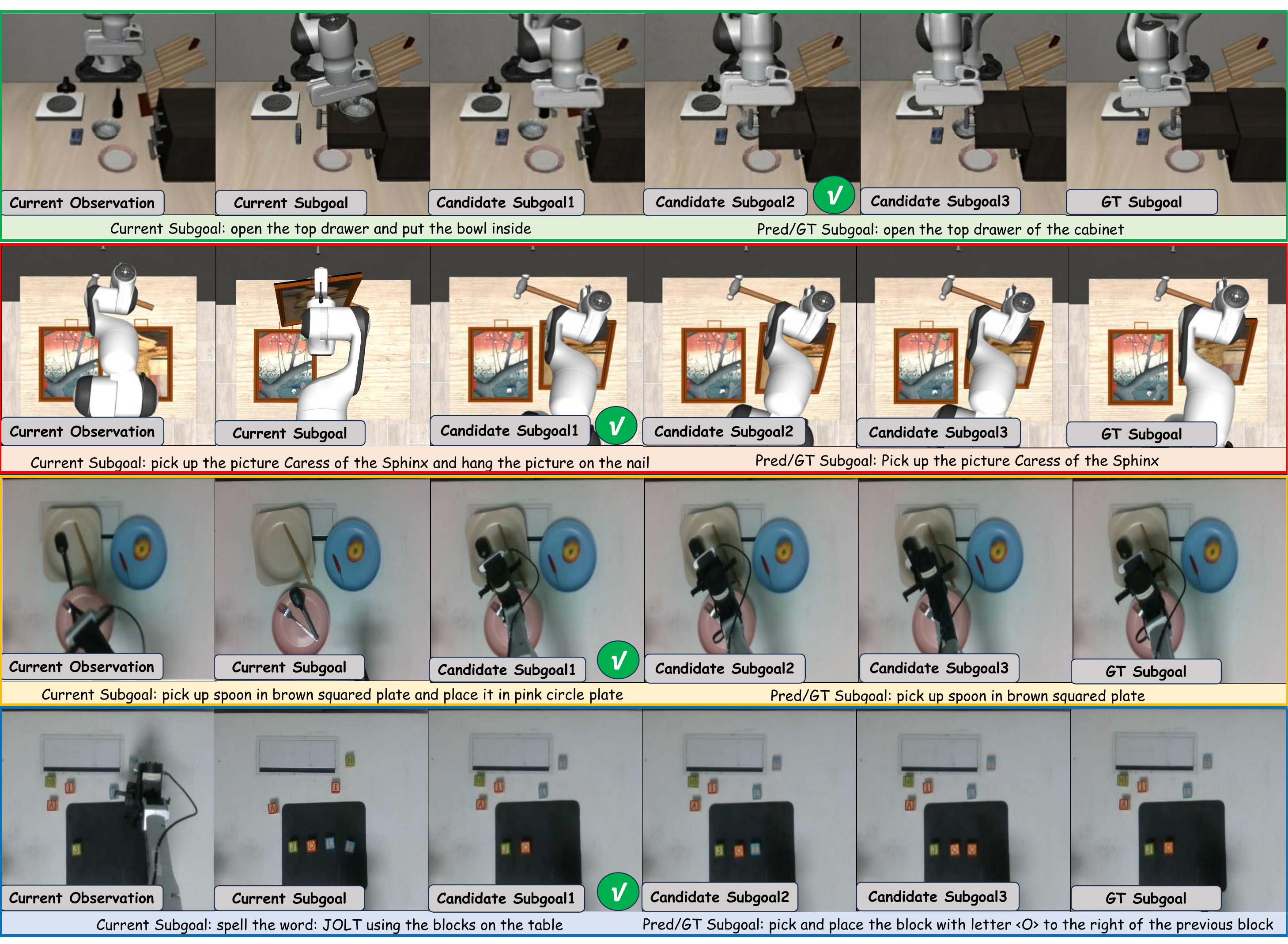}
  \caption{Qualitative visualization of generated subgoals.}
  \label{fig:bagel_quality}
  \vspace{-3mm}
\end{figure}

In this section, we evaluate the anticipation from two perspectives:
\begin{inparaenum}
    \item[1)] The subtask prediction accuracy of policy model $l_\theta$;
    \item[2)] The subgoal generation quality of dynamics model $P_\theta$.
\end{inparaenum}
To achieve this, we test anticipation models on 400 held-out samples per task. The qualitative visualizations are shown in Figure~\ref{fig:bagel_quality}, while the corresponding quantitative evaluation is presented in Table~\ref{tab:quality_result}. We observe that the anticipation model demonstrates impressive subtask prediction accuracy across various benchmarks. Regarding visual subgoal generation, the model demonstrates strong generalization to real-world tasks, maintaining robust performance even in complex scenarios. However, performance on simulated tasks is relatively lower, particularly for the VLABench benchmark. We attribute this gap to several factors: the simulated dataset being smaller, the challenge of generating intricate and unseen pictures, and the fact that the UMM was not pretrained on simulated data.
\begin{table}[htbp]
\centering
\caption{Quantitative results anticipation model.}
\begin{tabular}{lccccc}
\toprule
\multirow{2}{*}{\textbf{Benchmark}} & \multicolumn{1}{c}{\textbf{Textual Subgoal}} & \multicolumn{4}{c}{\textbf{Image Subgoal}} \\
\cmidrule(lr){2-2}\cmidrule(lr){3-6}
&\textbf{Pred. Acc.} &  \textbf{ PSNR} &\textbf{ MAE} & \textbf{SSIM} & \textbf{FID}\\
\midrule
\textbf{Libero}          & 84.4 & 20.4 &  9.4 & 0.85 & 31.0 \\
\textbf{VLABench}        & 88.8  & 15.5 & 19.0  & 0.76 & 55.1 \\
\textbf{Rearrange Objects} & 88.1  & 28.0 &  6.1 & 0.93 & 45.1 \\
\textbf{Spell Words}       & 98.9 & 26.4 &  6.9 & 0.92 & 34.7 \\
\bottomrule
\end{tabular}%
\label{tab:quality_result}
\vspace{-5mm}
\end{table}

\section{Related Works}

\textbf{Vision-Language-Action Models.} By leveraging the generalization ability from pretrained VLMs~\citep{steiner2024paligemma,bai2025qwen2}, VLA models have shown strong task performance across diverse manipulation tasks~\citep{kim2025fine,intelligence2025pi05visionlanguageactionmodelopenworld,bjorck2025gr00t}. A prominent line of work focuses on discretizing low-level actions and using the next token prediction framework to predict actions directly from multimodal inputs~\cite{octo_2023, kim2024openvla, pertsch2025fast, goyal2025vla}. Complementary approaches adopt continuous action representations through flow matching or diffusion-based techniques to better capture the fine-grained actions~\citep{black2024pi_0,wen2025dexvla,zhong2025dexgraspvla,deng2025graspvla,bjorck2025gr00t}. Efforts to further extend the generalization abilities of VLA models have focused on incorporating cross-modal perception and prediction abilities~\citep{zhen20243d,zhengtracevla,zhangdreamvla,cen2025worldvla,li2025unified,wang2025unified}. However, despite these advancements, VLA models still struggle with long-horizon tasks due to the compounding policy errors~\citep{fan2025long,zhang2025vlabench,gao2025vlaos,li2025simplevla}.

\textbf{Subgoal Generation for Long-Horizon Tasks.} To address compounding errors in long-horizon tasks, subgoal generation has been widely explored. A prominent research use VLMs to decompose tasks into natural language-based subtask~\citep{mu2023embodiedgpt,brohan2023can,zhou2025chatvla,zhou2505chatvla2,intelligence2025pi05visionlanguageactionmodelopenworld}. Meanwhile, several works direct generate visual-based subgoals via pretrained vision models to guide action generation~\citep{blackzero,bharadhwaj2024gen2act,wu2024robomind,nasiriany2024pivot,sun2025planning,zhao2025cot}. Additionally, multimodal approaches like VLA-OS~\citep{gao2025vlaos} and dVLA~\citep{wen2025dvla} integrate both textual and visual subgoals, enriching contextual understanding and improving policy robustness. Despite their advances, these methods often rely on fixed subgoal granularities, which may either overly fine-grained subgoals that introduce unnecessary complexity, or too sparse subgoal generation that fail to guide policy execution. Another line of works use implicit world modeling by predicting future observations~\citep{wu2023unleashing,cheang2024gr,li2025gr,li2025unified,bu2025univlalearningacttaskcentric,zhang2025upvlaunifiedunderstandingprediction,cen2025worldvla}. However, these tend to focus on single-step predictions, leading to overly fine-grained subgoals that lack of planning ability. In contrast, Anticipation-VLA dynamically adjusts subgoal granularity based on task progress, continuously refining subgoals to overcome these limitations.

\section{Conclusion}
In this paper, we propose Anticipation-VLA, a hierarchical framework with an adaptive, recursive subgoal generation mechanism to solve long-horizon embodied tasks. By dynamically refining multimodal subgoals based on execution progress, it reduces compounding errors and enables robust, generalizable policy execution. Empirical evaluation on both simulated and real-world tasks shows that our method significantly outperforms existing VLA baselines.

\textbf{Limitations.} Although Anticipation-VLA can generalize to unseen tasks, it still requires a few annotated subgoal demonstrations for finetuning. Additionally, despite our adaptive strategy significantly reducing subgoal generation frequency, visual subgoal generation remains computationally costly, causing occasional inference pauses. Future work could address these through improved pretraining for one-shot or zero-shot planning and more efficient inference via smaller models or acceleration techniques.
\section*{Impact Statement}
This paper presents work whose goal is to advance the field of Machine
Learning. There are many potential societal consequences of our work, none
of which we feel must be specifically highlighted here.


\bibliography{example_paper}
\bibliographystyle{icml2026}

\include{appendix}


\end{document}

%% file: appendix.tex
\newpage
\appendix
\onecolumn

\section{Computational Resources}
\label{app:compute}
All experiments were conducted using 4 NVIDIA H100 GPUs. Training the Anticipation Model on the collected dataset takes approximately 4–6 hours. Meanwhile, training the Goal-conditioned VLA typically completes in about 1–2 hours.

\section{Real-World Task Setup}
\subsection{Hardware Setup}
\label{app:hardware}
\begin{figure}[htbp]
    \centering
    \includegraphics[width=0.95\linewidth]{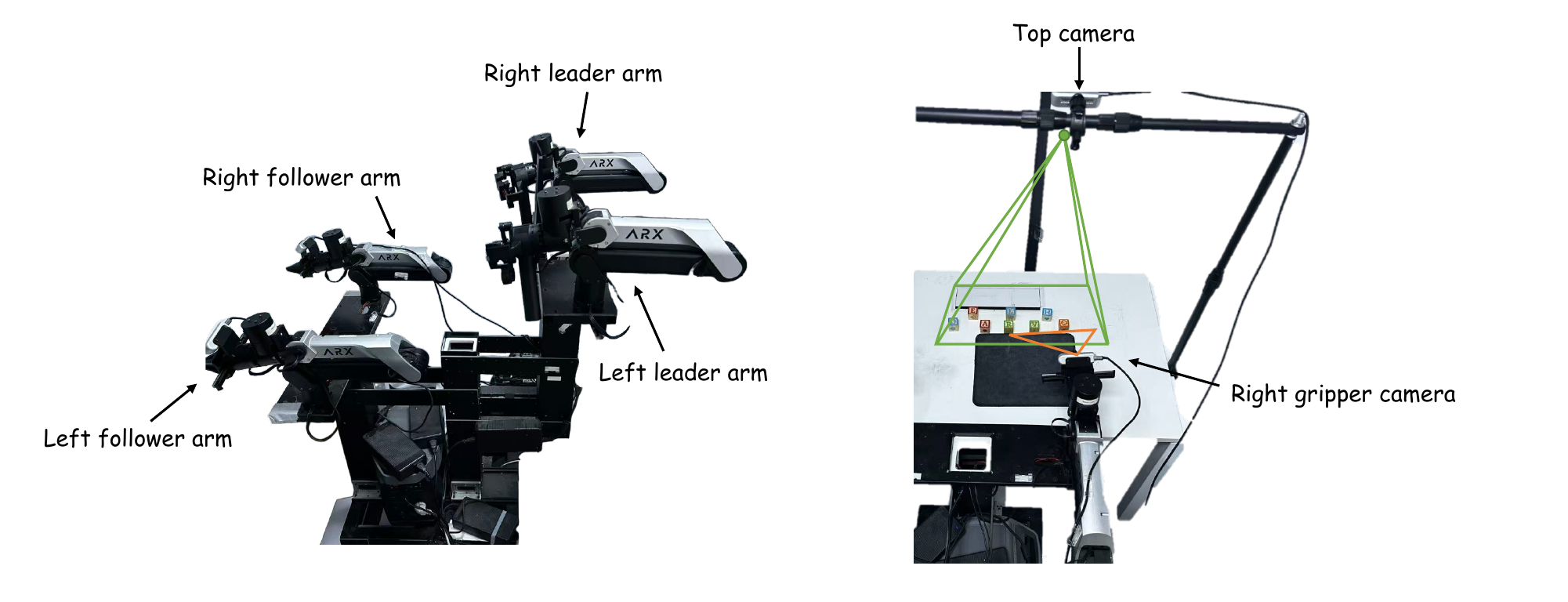}
    \caption{Hardware Illustration of ARX-X5 robotic platform.}
    \label{fig:hardware}
    \vspace{-3mm}
\end{figure}

We conduct real-world experiments on the Arx-X5 mobile manipulator platform (see Figure~\ref{fig:hardware}), a widely adopted system for training and benchmarking real-world robotic policies. We only use the right arm in our experiments. For visual image capturing, we use two Realsense D435i cameras. One of the camera is mounted at the top of the platform and the other one is on the robot's right arm. 

\subsection{Task Design}
\label{app:task}
We design two real-world manipulation tasks to evaluate the long-horizon execution for our framework.

\begin{enumerate}
\item \textbf{Rearrange Objects.} 
In this task, the robot is provided with a goal image that specifies the desired final configuration of the tabletop. The scene contains multiple everyday objects, including fruits (e.g., apples and lemons) and utensils (e.g., forks and knives), placed on plates of different shapes and colors. Given the target image, the robot is required to identify the objects to be manipulated and sequentially pick and place each object into its corresponding target location to match the goal configuration. Successful execution requires accurate visual grounding, object discrimination among multiple candidates, and reliable long-horizon manipulation under image-conditioned goals.

\item \textbf{Spell Words.} 
In this task, the robot receives a natural language instruction specifying a target word (e.g., “spell the word ICML”). A set of lettered blocks is scattered on the tabletop. The robot must correctly identify the blocks corresponding to the target letters and place them on the table in the correct left-to-right order to form the specified word. This task evaluates the model’s ability to ground language instructions into sequential object selection and ordered placement, as well as its robustness in long-horizon execution with language-conditioned goals.
\end{enumerate}

\subsection{Data Collection}
\label{app:data}
\begin{figure}[htbp]
    \centering
    \includegraphics[width=0.7\linewidth]{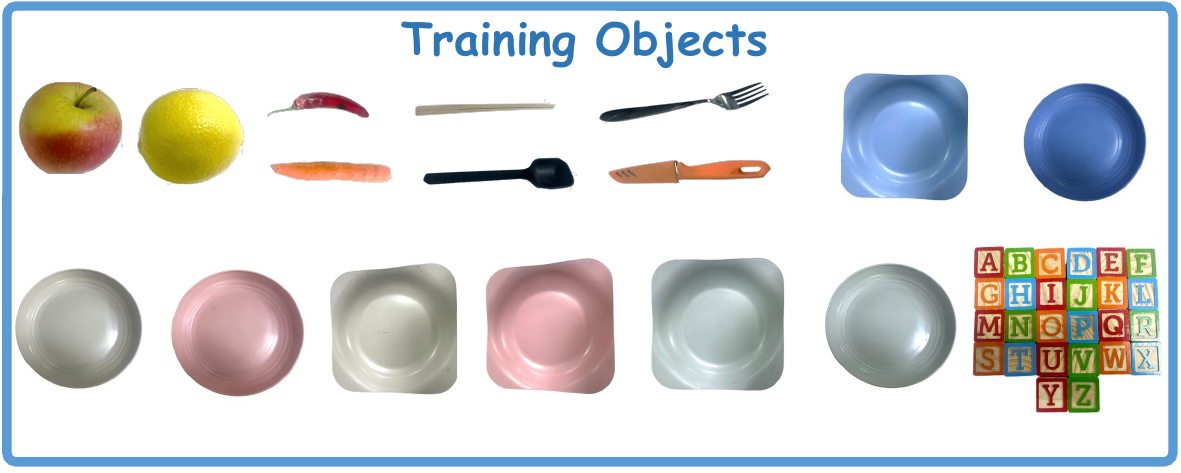}
    \caption{Seen objects for real-world tasks.}
    \label{fig:seen_obj}
    \vspace{-3mm}
\end{figure}

We collect expert demonstration data on the Arx-X5 robotic platform for both real-world tasks using human teleoperation. All the real-world objects for training can be found in Figure ~\ref{fig:seen_obj}.

For \textit{Rearrange Objects} task, we design 100 distinct scenes with varying object layouts and target configurations. For each scene, an expert operator provides a single successful demonstration, resulting in a total of 100 expert trajectories. These demonstrations cover diverse object arrangements and manipulation sequences, enabling the model to learn robust image-conditioned multi-object rearrangement behaviors.

For \textit{Spell Words} task, we construct 100 different scenes with varying distributions of lettered blocks and target words, where the word length ranges from three to five letters. Due to the increased task complexity and longer execution horizon, we collect 200 expert trajectories, with multiple demonstrations provided for each scene. This data set captures various execution strategies and ordering variations, supporting stable learning for language-conditioned sequential manipulation.

All demonstrations are recorded as multimodal trajectories, including visual observations and action sequences, and are used for training and evaluation in real-world experiments. 

\subsection{Generalization Evaluation}
\label{app:eval}
\begin{figure}[htbp]
    \centering
    \includegraphics[width=0.95\linewidth]{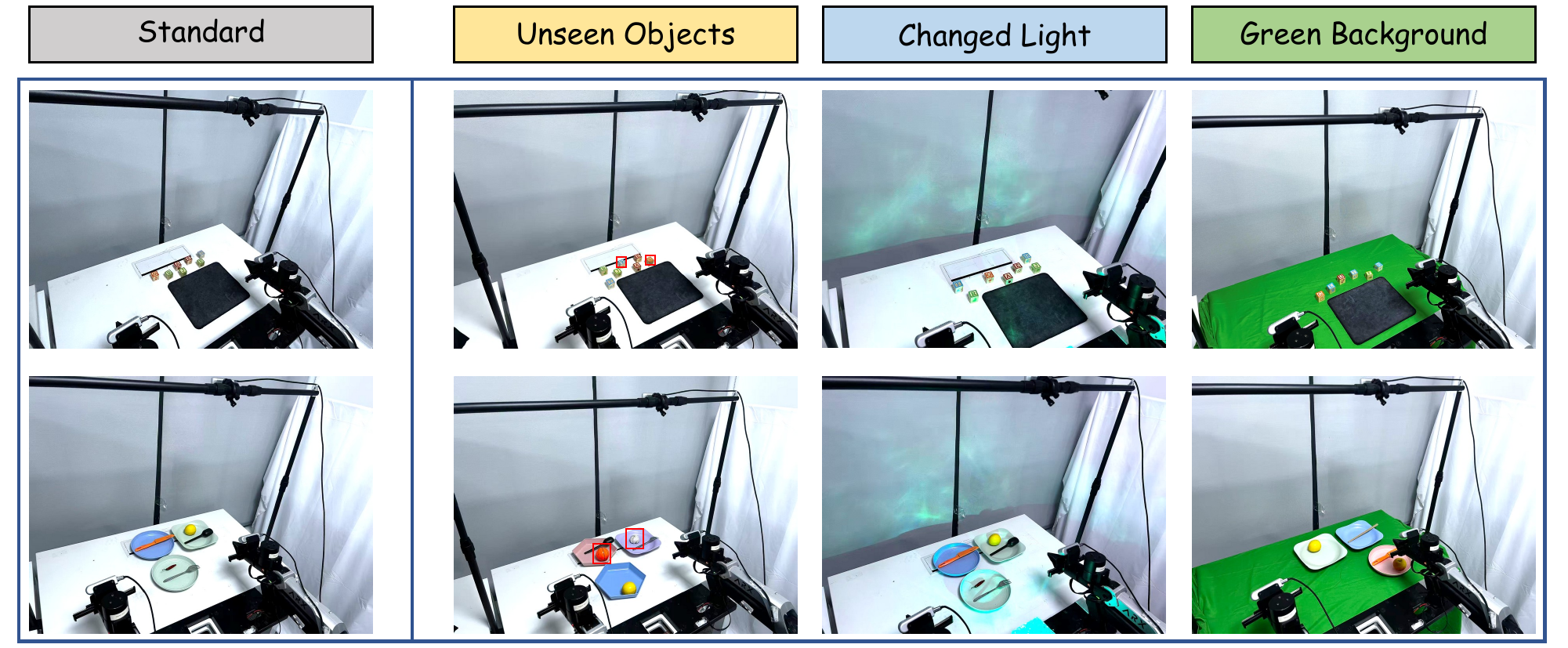}
    \caption{Unseen background for real-world tasks.}
    \label{fig:unseen_bcg}
    \vspace{-4mm}
\end{figure}

\begin{figure}[htbp]
    \centering
    \includegraphics[width=0.7\linewidth]{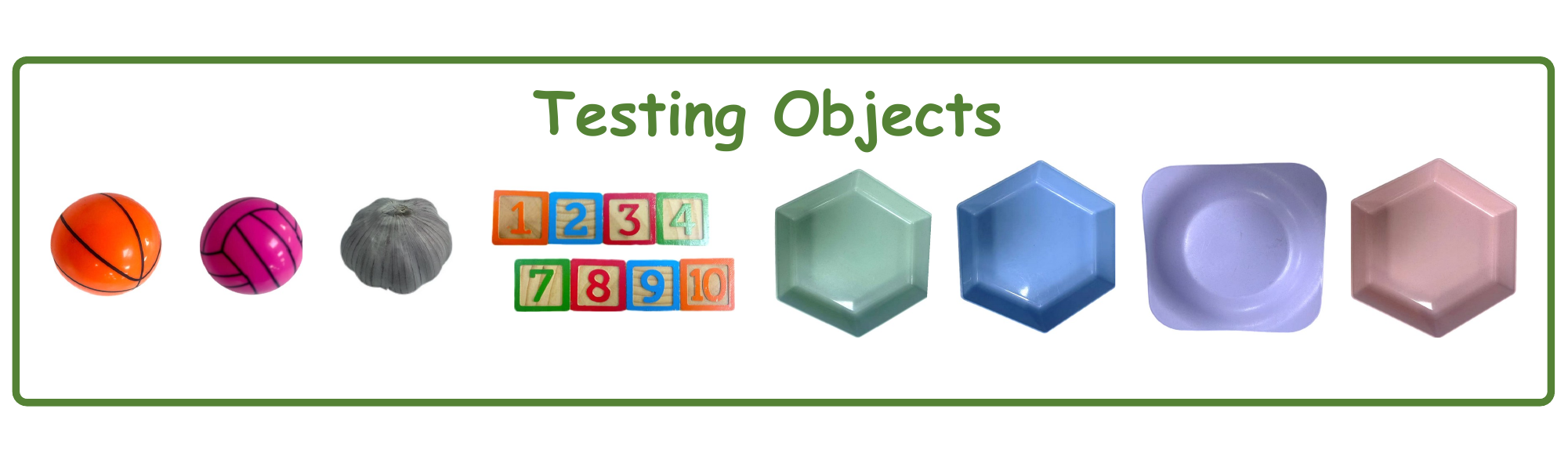}
    \caption{Uneen objects for real-world tasks.}
    \label{fig:unseen_obj}
    \vspace{-3mm}
\end{figure}

Beyond the standard evaluation on the held-out test set, we additionally conduct robustness tests under challenging conditions, including variations in environmental appearance (Figure~\ref{fig:unseen_bcg}) and scenarios involving unseen objects (Figure~\ref{fig:unseen_obj}). These settings intentionally introduce distribution shifts to evaluate the generalization of our model.

\section{Dataset Construction}
\label{app:data_construction}

\subsection{Hierarchical Subgoal Annotation} 
To train and evaluate the anticipation model with explicit hierarchical supervision, we annotate multi-level subgoals and corresponding language prompts for trajectories from both simulation benchmarks and real-world tasks.

\paragraph{Subgoal Hierarchy Definition.}
For each trajectory, we define a hierarchical subgoal structure with multiple levels of granularity.
Let level $h=0$ denote the highest-level task goal (i.e., the original instruction or final target), and larger values of $h$ correspond to increasingly fine-grained subgoals until reaching atomic, directly executable objectives.

Each subgoal at level $h$ is represented as a multimodal goal $g_h = (\ell_h, s_h)$,
where $\ell_h$ denotes a language subgoal instruction and $s_h$ denotes a visual target image specifying the desired future state. 

\paragraph{Annotation Protocol.}
The annotation process follows a top-down and recursive decomposition strategy:
\begin{enumerate}
    \item \textbf{High-level goal identification.}  
    The original task instruction or final goal image is assigned as the level-0 goal.
    \item \textbf{Recursive subgoal decomposition.}  
     Annotators iteratively decompose each goal into a sequence of intermediate subgoals that are achievable from the current state and correspond to meaningful progress toward the final goal.
    \item \textbf{Multimodal labeling.}  
    For each subgoal, annotators provide a language prompt describing the intended subgoal (e.g., ``pick up the block with letter \textless C\textgreater'') together with a visual target image selected from the trajectory or rendered from the simulator.
    \item \textbf{Granularity consistency.}  
    Subgoals at the same hierarchy level across different trajectories are annotated to maintain consistent semantic abstraction and comparable planning granularity.
\end{enumerate}

\paragraph{Annotated Datasets.} Our annotated dataset is composed of four distinct parts:

\begin{itemize}
    \item \textbf{LIBERO (Simulation)}: 40 expert trajectories sampled from the LIBERO official expert dataset.
    \item \textbf{VLABench (Simulation)}: 100 expert trajectories collected from the VLABench simulator.
    \item \textbf{Pick-and-Place (Real World)}: 100 expert trajectories collected via human teleoperation.
    \item \textbf{Spell Words (Real World)}: 200 expert trajectories collected via human teleoperation.
\end{itemize}

\subsection{Anticipation Dataset Preparation.}
\label{app:ant_data}
With the subgoal sequences at hand, we convert them into concrete supervised learning signals. In this process, we construct four datasets, which are trained jointly within a unified multimodal model (UMM). Below are the specific methods used to construct each dataset:

\begin{itemize}
    \item \textbf{Dynamics Dataset:} This dataset is built using a systematic sampling strategy. For each subgoal \( g \), we randomly select a frame \( f \) from the interval between \( g \) and its preceding subgoal at the same level. The model takes \( f.\text{image} \) (the initial observation) and \( g.\text{action} \) (the target action) as inputs, with the visual state \( g.\text{image} \) as the prediction target.
    
    \item \textbf{Policy Dataset:} To construct this dataset, we sample a frame \( f \) from between the frames of two consecutive subgoals \( (g_{\text{prev}}, g) \) in \( \mathcal{S} \). For input, we use \( f.\text{image} \) as the current observation, along with \( g.\text{parent}.\text{image} \) and \( g.\text{parent}.\text{action} \) as goal observations. The prediction target is \( g.\text{action} \).
    
    \item \textbf{Inverse Dynamics Dataset:} This dataset is constructed similarly to the Dynamics Dataset, where we sample a frame \( f \) between subgoals. However, here, the input consists of \( f.\text{image} \) and \( g.\text{image} \), and the target is \( g.\text{action} \).
    
    \item \textbf{Optimal Value Dataset:} To construct this dataset, we first sample a frame \( f_1 \) for each subgoal \( g \) using the same method as the Dynamics Dataset. Next, we sample another frame \( f_2 \) in the range of \( (f_1.\text{frame} + \delta, g.\text{frame} + \epsilon) \) to serve as the current observation. We then label the progress toward subgoal \( g \) made by \( f_2 \) relative to \( f_1 \) using the following classification:

    \[
    \text{label}(f_1, f_2, g) =
    \begin{cases}
        \text{Insufficient Progress} & \text{if } f_2.\text{frame} \in [0, f_1.\text{frame} + \beta) \\
        \text{Sufficient Progress} & \text{if } f_2.\text{frame} \in [f_1.\text{frame} + \beta, g.\text{frame} - \gamma) \\
        \text{Goal Achievement} & \text{if } f_2.\text{frame} \in [g.\text{frame} - \gamma, \infty)
    \end{cases}
    \]
\end{itemize}

In this setup, the labels categorize the frame transitions as "No Progress," "Progress," or "Achieved," based on the temporal differences and the progress made toward the goal.

\paragraph{Task-Specific Characteristics.}
For Libero and VLABench, subgoals typically correspond to object-centric or spatial manipulation primitives (e.g., opening drawers, picking objects, or placing objects).
For the real-world Rearrange Objects task, subgoals emphasize precise object selection and placement under visual variability.
For the Spell Words task, subgoals reflect sequential symbolic reasoning, where each subgoal corresponds to selecting and placing a specific letter block in the correct order. The annotated target words range from 3 to 5 letters, resulting in diverse task horizons.

\section{Implementation Details}
\label{app:impl}
\subsection{Inference Procedure}
\begin{algorithm}
\caption{Anticipation-VLA Inference Procedure}
\label{alg:anticipation_vla_inference}
\begin{algorithmic}[1]
\REQUIRE Current state $s_0$, Final goal $g_\text{final}$
\STATE \textbf{Models:} Optimal Value Model $V^\ast$, Anticipation Model $G$, Goal-conditioned Policy Model $\pi$
\STATE \textbf{Parameters:} Planning Check Interval $K$, Goal Achievement Threshold $\delta$, Stack Max Depth $d$
\STATE Initialize  $\mathcal{G}_{\text{stack}} \leftarrow \emptyset$, $s \leftarrow s_0$, $s_\text{pref}\leftarrow s_0$, $t\leftarrow0$

\STATE $\mathcal{G}_{\text{stack}}.\text{push}(g_\text{final})$ \hfill Push the final goal
\WHILE{$\mathcal{G}_{\text{stack}}$ is not empty}
\STATE $g \leftarrow \mathcal{G}_{\text{stack}}.\text{peek}()$ \hfill Peek the active goal
\STATE $a \leftarrow \pi(\cdot | s, g)$ \hfill Generate low-level action
\STATE $s' \leftarrow P(\cdot|s, a)$ \hfill Execute action and observe new state
\STATE $s \leftarrow s'$ \hfill Update state for next timestep
\STATE $t \leftarrow t+1$ \hfill Increment timestep

\IF{$t \bmod K = 0$ \text{and} $t > 0$}  
\STATE \textbf{if} $|V^\ast(s, g)-V^\ast(s_g, g)| < \delta$ \textbf{then} \hfill \textcolor{DarkBlue}{Condition 1: Goal Achieved (Pop)}
\STATE \quad $\mathcal{G}_{\text{stack}}.\text{pop}()$ \hfill Pop achieved subgoal
\STATE \textbf{endif}
\STATE \textbf{else if} {$|V^\ast(s, g) - V^\ast(s_\text{prev}, g)| < \delta$ and $|\mathcal{G}_\text{stack}| < d$} \textbf{then} \hfill \textcolor{orange}{Condition 2: Insufficient Progress (Refine)}
\STATE \quad  $g' \leftarrow  G(\cdot|s, g)$ \hfill Generate refined subgoal
\STATE \quad $\mathcal{G}_{\text{stack}}.\text{push}(g')$ \hfill Push the refined subgoal
\STATE \textbf{endif}
\STATE \textbf{else if} {$|V^\ast(s, g) - V^\ast(s_\text{prev}, g)| < \delta$ and $|\mathcal{G}_\text{stack}| = d$} \textbf{then} \hfill \textcolor{DarkRed}{Condition 3: Local Stagnate (Backtrack)}
\STATE \quad  $\mathcal{G}_{\text{stack}} \leftarrow \emptyset$ \hfill Clear goal stack
\STATE \quad $\mathcal{G}_{\text{stack}}.\text{push}(g_\text{final})$ \hfill Push the final goal
\STATE \quad  $s\leftarrow s_0$ \hfill Reset the robot pose
\STATE \textbf{endif}
\STATE \textbf{endif}
\STATE $s_\text{prev} \leftarrow s$ \hfill Update state $s_\text{prev}$
\ENDIF
\ENDWHILE
\end{algorithmic}
\end{algorithm}
The inference Procedure of Anticipation Model is described in algorithm ~\ref{alg:anticipation_vla_inference}. We maintain a stack-driven high-level planner, where each element on the stack represents a subgoal. The low-level VLA always receives the top element of the stack as the current subgoal to be completed. The high-level planner uses the optimal value model to check the progress to complete current subgoal $g$ after the low-level VLA executes for $K$ steps. If there has been almost no progress since the last observation $s_{\text{prev}}$, the high-level planner will generate a finer-grained subgoal $g_{\text{new}}$ that completes before the current subgoal $g$, based on $s$ and $g$. If $g$ has already been completed, the planner will pop it from the stack, so that the next, coarser-grained subgoal will be at the top again. If $s$ does achieve some progress compared to $s_{\text{prev}}$ but $g$ is still not achieved, we will continue using $g$ as current subgoal. 

The stack-based subgoal maintenance ensures the top subgoal starts after the bottom subgoal begins and finishes before it completes, enabling finer grained decomposition and tighter synchronization with low-level VLA's execution.

\subsection{Anticipation Model}
\begin{figure}[htbp]
    \centering
    \includegraphics[width=0.85\linewidth]{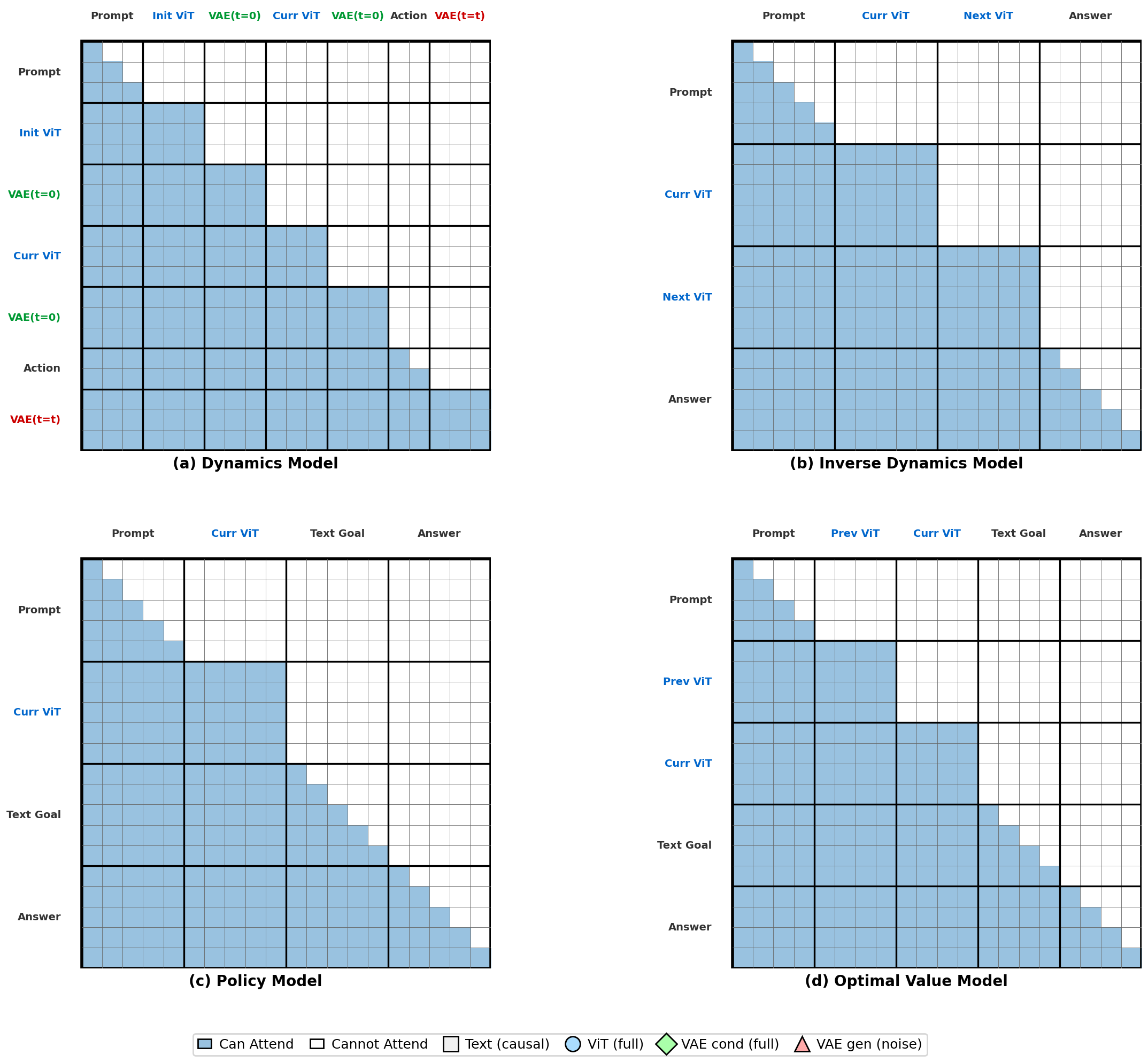}
    \caption{\textbf{Causal mask configurations for four anticipation tasks.} 
    (a) Dynamics Model sequence comprises prompt tokens (causal), initial ViT features (full), VAE condition tokens at t=0 (full), current ViT features (full), action tokens (causal), and VAE generation tokens at t=t (noise). 
    (b) Inverse Dynamics Model sequence comprises Prompt (causal), current ViT (full), next ViT (full), and answer tokens (causal). Both ViT blocks apply bidirectional attention internally.
    (c) Policy Model sequence compromises Prompt (causal), current ViT (full), text goal (casual), and answer (causal).
    (d) Optimal Value Model sequence compromises Prompt (causal), previous ViT (full), current ViT (full), text goal (causal), and answer (causal).}
    \label{fig:casual_mask}
\end{figure}

We finetuned Bagel as the anticipation model and the value model. Here, we provide the casual mask of the anticipation model during training, as illustrated in Figure~\ref{fig:casual_mask}.

The training and inference hyperparameters of the anticipation model are provided in Table~\ref{tab:hyper_anticipation}.

\begin{table}[htbp]
\centering
\caption{Hyperparameters for Anticipation Model.}
\begin{tabular}{l c}
\toprule
\textbf{Parameters} & \textbf{Value} \\
\midrule
lr & $2\mathrm{e}{-5}$ \\
Cross entropy weight & 0.01 \\
MSE weight & 1.0 \\
Max latent size & 64 \\
Timestep shift & 4 \\
Cfg interval & [0.4, 1.0] \\
Text scale & 6 \\
Image scale & 2 \\
\bottomrule
\end{tabular}
\label{tab:hyper_anticipation}
\end{table}

\subsection{Goal-conditioned VLA Model}
We instantiate our policy using $\pi_{0.5}$ ~\cite{intelligence2025pi05visionlanguageactionmodelopenworld}, adapting it to condition on anticipated subgoals. The input sequence is constructed by concatenating multi-view observations $\mathbf{s}_o^t$, the subgoal image $s_g^t$, robot state $\mathbf{q}$, and subgoal instruction $\ell_g^t$. The model learns the trajectory distribution $\pi_\theta(\mathbf{a}^{t:t+h}|\mathbf{s}_o^t, g_t)$, conditioned on the goal tuple $g_t = (s_g, \ell_g)$. To bridge the domain gap between ground-truth and generated subgoals, we apply a robustness strategy where tokens in the goal image $s_g$ are randomly masked with a probability $p$ during training. Table~\ref{tab:goal-conditioned VLA} lists hyperparameters for goal-conditioned $\pi_{0.5}$ fine-tuning. We finetune VLAs until convergence following the official implementation \url{https://github.com/Physical-Intelligence/openpi}.

\begin{table}[H]
    \centering
    \caption{Hyperparameters of Goal-conditioned VLA}
    \begin{tabular}{ll}
        \toprule
        \textbf{hyperparameter} & \textbf{value} \\
        \midrule
        \# GPUs & 4 $\times$ NVIDIA H100 (80GB VRAM)  \\

        learning rate (LR) 
        &  2.5e-5 peak LR (1K steps linear warmup, 29K steps cosine decay to 2.5e-6)\\

        total batch size 
        & 64 \\

\# train steps & 5k for LIBERO \\
 & 10k for VLABench \\
 & 10k for real-world tasks \\

 input images 
 & 1 head camera image, 1 wrist camera image for LIBERO \\
 & 1 head camera image, 1 wrist camera image for VLABench \\
 & 1 head camera image, 1 right camera image real-world tasks \\

        input image size 
        &  224 × 224 px \\

        use observation history 
        & no (use single-step inputs)  \\

action chunk size 
& 10 steps for LIBERO  \\
& 10 steps for VLABench \\
& 20 steps for real-world tasks \\

        \bottomrule
    \end{tabular}
    \label{tab:goal-conditioned VLA}
\end{table}

\section{Prompts Details}
\label{app:prompt}
\subsection{Prompt for Libero}
\begin{tcolorbox}[
    colback=gray!20,
    colframe=gray!70,
    sharp corners,
    boxrule=0.5pt,
    title=Prompt for Libero,
    fonttitle=\bfseries,
    enhanced jigsaw,
    breakable,             
    left=2mm, right=2mm,   
    top=1mm, bottom=1mm,
    width=\linewidth       
]
\ttfamily 
\obeylines 
\textbf{Policy Model}
You are a vision-language model with advanced reasoning abilities.
Your task is to carefully observe the current state image and the goal (text and image) of a given libero task, and then **predict the next action to better achieve this goal**.

\#\#\# Environment description:
You are in a robot manipulation workspace that contains:
- Various household objects: plates, mugs, bowls, food items, kitchen appliances
- Manipulable items: drawers, cabinets, microwave, stove, trash bins
- A robotic arm with a gripper for manipulation

\#\#\# Action space:
- "pick up the [object] and place it on/in [location]"
- "pick up [object]"
- "place it on/in [location]"
- "grasp [object]"
- "open/close [object]"
- "turn on [object]"

\#\#\# Task:
You will receive:
1. **Current observation**: An image showing the robot's current view
2. **Current goal description**: A text description of the current goal to achieve
3. **Current goal observation**: An image of the current goal to achieve (may be absent for high-level tasks)

Your task is to predict what action the robot should take from the current state to progress toward the current goal.
You should output the action directly. Do not include any other output.

\par\medskip
\textbf{Inverse Dynamics Model}
You are a vision-language model with advanced reasoning abilities.
Your task is to carefully observe the current state image, next state image, and then **infer the action taken from the current state to reach the next state**.

\#\#\# Environment description:
You are in a robot manipulation workspace that contains:
- Various household objects: plates, mugs, bowls, food items, kitchen appliances
- Manipulable items: drawers, cabinets, microwave, stove, trash bins
- A robotic arm with a gripper for manipulation

\#\#\# Action space:
- "pick up the [object] and place it on/in [location]"
- "pick up [object]"
- "place it on/in [location]"
- "grasp [object]"
- "open/close [object]"
- "turn on [object]"

\#\#\# Task:
You will receive:
1. **Current observation**: An image (first image) showing the current state
2. **Next observation**: An image (second image) showing the next state to reach

Your task is to determine the action that was most likely taken to transition from the current observation to the next observation.
You should output the action directly. Do not include any other output.
\par\medskip
\textbf{Dynamics Model}
You are now acting as a **world model** that simulates robot manipulation task execution.
Your task is to predict the **next state observation after executing the action at the current state**.

\#\#\# Environment description:
- You are observing a robot workspace with various objects that can be manipulated
- The environment is from the LIBERO dataset, containing realistic household manipulation tasks
- Common objects include: plates, mugs, bowls, food items, kitchen appliances, drawers, etc.
- Tasks involve pick-and-place, opening/closing, and spatial arrangement operations

\#\#\# Action space:
- "pick up the [object] and place it on/in [location]"
- "pick up [object]"
- "place it on/in [location]"
- "grasp [object]"
- "open/close [object]"
- "turn on [object]"

\#\#\# Task:
You will receive:
1. **Init observation**: An image showing the initial state
2. **Current observation**: An image showing the robot's current view
3. **Action**: A text describing the manipulation to execute

Your task is to predict the **next subgoal frame of visual observation after executing the action at the current state**.

\#\#\# Important notes:
- Maintain **visual coherence** of the scene
- Accurately simulate the physical effect of the manipulation action
- Keep object positions and states consistent with the action description
- Preserve the background and unaffected objects
- Show appropriate changes in gripper position and object states

\par\medskip
\textbf{Optimal Value Model}
You are a vision-language model with advanced reasoning abilities.
Your task is to carefully observe two images from a robot manipulation task along with textual goal descriptions, and then **evaluate the progress status**.

\#\#\# Environment description:
You are in a robot manipulation workspace that contains:
- Various household objects: plates, mugs, bowls, food items, kitchen appliances
- Manipulable items: drawers, cabinets, microwave, stove, trash bins
- A robotic arm with a gripper for manipulation

\#\#\# Action space:
- "pick up the [object] and place it on/in [location]"
- "pick up [object]"
- "place it on/in [location]"
- "grasp [object]"
- "open/close [object]"
- "turn on [object]"

\#\#\# Task:
You will receive:
1. **Previous observation**: An image (first image) showing the robot's previous state
2. **Current observation**: An image (first image) showing the robot's current state
3. **Goal description**: A text description of the goal to achieve

Your task is to evaluate the progress status by comparing the two frames and the textual goal descriptions, then classify into one of three categories:

**Category 0 (No Progress/Stagnant)**:
- The current state shows MINIMAL or NO progress from the previous state toward the subgoal
- The change between previous and current is negligible
- The subgoal is still far from being achieved
- Examples: slight camera jitter, minimal gripper movement, waiting state, preparation phase, action not yet started

**Category 1 (Progress/Forward but Not Yet Complete)**:
- The current state shows SIGNIFICANT progress from the previous state
- Clear advancement toward the subgoal is visible
- But the subgoal is NOT yet achieved - the robot is still in the process of completing it
- Examples: successfully grasped object, moved closer to target location, mid-way through placing action, drawer partially opened

**Category 2 (Achieved/Completed)**:
- The current state shows that the subgoal has been ACHIEVED or is VERY CLOSE to completion
- The described subgoal action appears completed
- Examples: object successfully placed on target, door fully closed, grasp completed, target location reached

\#\#\# Output specification:
You should output **only a single integer** representing the progress category:
- **0**: No progress (stagnant or minimal change)
- **1**: Progress (significant forward movement but goal not reached)
- **2**: Achieved (goal reached or nearly reached)

**Do NOT include any explanation, reasoning, or additional text**
**Only output the integer number itself (0, 1, or 2)**

\#\#\# Evaluation tips:
- Compare object positions, gripper states, and scene configurations between the two images (previous and current)
- Consider the temporal direction: previous → current (how much progress was made?)
- Match the visual changes to the subgoal description: does the current state align with achieving the subgoal?
- Look for visual cues: grasping success, object placement, spatial relationships, gripper state changes
- Consider action completion: has the described action in the subgoal been executed successfully?

\end{tcolorbox}

\subsection{Prompt for VLABench}
\begin{tcolorbox}[
    colback=gray!20,
    colframe=gray!70,
    sharp corners,
    boxrule=0.5pt,
    title=Prompt for VLABench,
    fonttitle=\bfseries,
    enhanced jigsaw,
    breakable,             
    left=2mm, right=2mm,   
    top=1mm, bottom=1mm,
    width=\linewidth       
]
\ttfamily 
\obeylines 
\textbf{Policy Model}
You are a vision-language model with advanced reasoning abilities.
Your task is to observe the goal, and then **describe the current state and give the next action to achieve the goal**.

\#\#\# Environment description:
You are in a robot manipulation workspace that contains:
- Various household objects: plates, mugs, bowls, food items, kitchen appliances
- Manipulable items: drawers, cabinets, microwave, stove, trash bins
- A robotic arm with a gripper for manipulation

\#\#\# Action space:
Your actions should be in natural language format, such as:
- "grasp the [object]"
- "move to the [location/object]"
- "place the [object] on the [location]"
- "put the [object] on/in the [location]"
- "open/close the [object]"
- "turn on/off the [object]"
- "release the [object]"
- "push the [object] [direction/location]"

\#\#\# Task:
Given the current observation and the overall goal, determine what the robot should do next.You should output the next action directly. Do not include any other output.
\par\medskip
\textbf{Dynamics Model}
You are now acting as a **world model** that simulates robot manipulation task execution.
Your task is to predict the **next frame of visual observation**, given the following inputs:
- A **current observation image** showing the robot's view of the workspace
- An **action description** describing the manipulation to execute

\#\#\# Environment description:
- You are observing a robot workspace with various objects that can be manipulated
- The environment is from the LIBERO dataset, containing realistic household manipulation tasks
- Common objects include: plates, mugs, bowls, food items, kitchen appliances, drawers, etc.
- Tasks involve pick-and-place, opening/closing, and spatial arrangement operations

\#\#\# Action space:
Actions are described in natural language, such as:
- "grasp the [object]"
- "move to the [location/object]"
- "place the [object] on the [location]"
- "put the [object] on the [location]"
- "open/close the [object]"
- "turn on the [object]"
- "get close to the [object]"
- "push the [object] [direction/location]"
- "release the [object]"

Your task is to **predict the next image** that results from executing the given action from the current observation.

You must:
- Maintain **visual coherence** of the scene
- Accurately simulate the physical effect of the manipulation action
- Keep object positions and states consistent with the action description
- Preserve the background and unaffected objects
- Show appropriate changes in gripper position and object states

\par\medskip
\textbf{Inverse Dynamics Model}
You are a vision-language model with advanced reasoning abilities.
Your task is to carefully observe the current state image and goal state image, and then **describe the current state and goal state, and give the action taken in the current state to achieve the goal**.

\#\#\# Environment description:
You are in a robot manipulation workspace that contains:
- Various household objects: plates, mugs, bowls, food items, kitchen appliances
- Manipulable items: drawers, cabinets, microwave, stove, trash bins
- A robotic arm with a gripper for manipulation

\#\#\# Action space:
Your actions should be in natural language format, such as:
- "grasp the [object]"
- "move to the [location/object]"
- "place the [object] on the [location]"
- "put the [object] on/in the [location]"
- "open/close the [object]"
- "turn on/off the [object]"
- "release the [object]"
- "push the [object] [direction/location]"

\#\#\# Task:
Given the current observation and the goal observation, determine the action the model most probably does to achieve the goal state.

You should output the corresponding action directly. Do not include any other output.

\par\medskip
\textbf{Optimal Value Model}
You are a vision-language model with advanced reasoning abilities.
Your task is to carefully observe three images from a robot manipulation task, and then **evaluate the progress status**.

\#\#\# Environment description:
You are in a robot manipulation workspace that contains:
- Various household objects: plates, mugs, bowls, food items, kitchen appliances
- Manipulable items: drawers, cabinets, microwave, stove, trash bins
- A robotic arm with a gripper for manipulation

\#\#\# Task context:
You will be given three frames from the same task execution:
- **Image 1 (Previous)**: The state at an earlier time step
- **Image 2 (Current)**: The current state
- **Image 3 (Goal)**: The target state to achieve

All three images are part of **subtasks within the same overarching task framework**.

\#\#\# Action space:
The robot performs actions in natural language format, such as:
- "grasp the [object]"
- "move to the [location/object]"
- "place the [object] on the [location]"
- "put the [object] on/in the [location]"
- "open/close the [object]"
- "turn on/off the [object]"
- "release the [object]"
- "push the [object] [direction/location]"

\#\#\# Task:
Evaluate the progress status by comparing the three frames and classify into one of four categories:

**Category 0 (Regression/Backward)**:
- The current state has moved BACKWARD compared to the previous state
- The robot appears to have regressed or moved away from the goal
- Examples: object dropped back, incorrect action executed, further from target

**Category 1 (No Progress/Stagnant)**:
- The current state shows MINIMAL or NO progress from the previous state
- The change between previous and current is negligible (< a small threshold)
- Examples: slight camera jitter, minimal gripper movement, waiting state

**Category 2 (Progress/Forward)**:
- The current state shows SIGNIFICANT progress from the previous state
- Clear advancement toward the goal is visible
- The change is substantial ($\geq$ a meaningful threshold)
- But the goal is NOT yet reached
- Examples: successfully grasped object, moved closer to target location

**Category 3 (Completed/Near-Completion)**:
- The current state is AT or VERY CLOSE TO the goal state
- The task appears completed or almost completed
- The current state has finished the goal and is executing next task (For this task, you need to predict the success state of the goal)
- Examples: object placed on target, door fully closed, goal configuration achieved

\#\#\# Output specification:
You should output **only a single integer** representing the progress category:
- **0**: Regression (moved backward)
- **1**: No progress (stagnant or minimal change)
- **2**: Progress (significant forward movement)
- **3**: Completed (goal reached or nearly reached)

**Do NOT include any explanation, reasoning, or additional text**
**Only output the integer number itself (0, 1, 2, or 3)**

\#\#\# Evaluation tips:
- Compare object positions, gripper states, and scene configurations across all three images
- Consider the temporal direction: previous $\rightarrow$ current (did we move forward or backward?)
- Assess proximity to goal: current v.s. goal (how close are we?)
- Look for visual cues: grasping success, object placement, spatial relationships
\end{tcolorbox}
\subsection{Prompt for Rearrange Objects}
\begin{tcolorbox}[
    colback=gray!20,
    colframe=gray!70,
    sharp corners,
    boxrule=0.5pt,
    title=Prompt for Rearrange Objects,
    fonttitle=\bfseries,
    enhanced jigsaw,
    breakable,             
    left=2mm, right=2mm,   
    top=1mm, bottom=1mm,
    width=\linewidth       
]
\ttfamily 
\obeylines 
\textbf{Policy Model}
You are a robot learning agent. Given the current observation and goal information, predict the next action to take.

For level 2 tasks (atomic actions):
- You will receive the current observation image and a goal description with goal observation image
- Predict the immediate next action (e.g., "pick up apple in pink circle plate", "place it in brown squared plate")

For level 1 tasks (composite actions):
- You will receive the current observation image and goal observation image (final state)
- Goal description will be: (No goal description for high-level task)
- Predict the composite action to achieve the goal state

Action space includes:
- pick up [object] in [location]
- place it in [location]
- [object] can be: apple, pepper, carrot, lemon, fork, knife, spoon, chopstick
- [location] can be: [color] [shape] plate (e.g., pink circle plate, brown squared plate)

Output only the action description, nothing else.
\par\medskip
\textbf{Dynamics Model}
You are a robot dynamics model. Given the current state and an action, predict the next state after executing the action.

Input:
- Initial observation image (starting state)
- Current observation image
- Action to execute
- (For level 1 tasks) Goal observation image (final target state)

Output:
- The observation image after executing the action

The action space includes pick and place operations on various objects (apple, pepper, carrot, lemon, fork, knife, spoon, chopstick) between different colored plates (pink, brown, blue, green circle/squared plates).

Note: The model gives higher weight to samples near subgoals to better capture critical transitions.
\par\medskip
\textbf{Inverse Dynamics Model}
You are a robot inverse dynamics model. Given two consecutive observations, infer what action was taken between them.

Input:
- Current observation image
- Next observation image

Output:
- The action that was executed to transition from current to next state

Action space includes:
- pick up [object] in [location]
- place it in [location]
- [object] can be: apple, pepper, carrot, lemon, fork, knife, spoon, chopstick
- [location] can be: [color] [shape] plate

Output only the action description, nothing else.

\par\medskip
\textbf{Optimal Value Model}
You are a robot progress evaluator. Given previous and current observations along with a goal, evaluate the progress towards achieving the goal.

For level 2 tasks:
- Goal is specified as a text description (e.g., "pick up apple in pink circle plate")

For level 1 tasks:
- Goal description is: (No goal description for high-level task)
- Goal observation image shows the target final state

Output:
- 0: No progress (current state is similar to previous state)
- 1: Progress (moved closer to the goal but not yet achieved)
- 2: Achieved (goal has been accomplished)

Note: The dataset includes extra samples near subgoals to better capture difficult "no progress" cases where the robot is close to achieving a goal but hasn't made the final step yet.

Output only the number (0, 1, or 2), nothing else.
\end{tcolorbox}
\subsection{Prompt for Spell Words}
\begin{tcolorbox}[
    colback=gray!20,
    colframe=gray!70,
    sharp corners,
    boxrule=0.5pt,
    title=Prompt for Spell Words,
    fonttitle=\bfseries,
    enhanced jigsaw,
    breakable,             
    left=2mm, right=2mm,   
    top=1mm, bottom=1mm,
    width=\linewidth       
]
\ttfamily 
\obeylines 
\textbf{Policy Model}
You are a robot learning agent. Given the current observation and goal information, predict the next action to take.

For level 2 tasks (atomic actions):
- You will receive the current observation image and a goal description with goal observation image
- Predict the immediate next action (e.g., "pick up the block with letter: <A>", "place the block on the table")

For level 1 tasks (composite actions):
- You will receive the current observation image and overall task description
- Goal description will be the overall task (e.g., "spell the word: CAT using the blocks on the table")
- Predict the composite action to achieve the current sub-goal

Action space includes:
- pick up the block with letter: <LETTER>
- place the block on the table
- place the current block to the right of the previous block
- place the block with letter <LETTER> on the table
- place the block with letter <LETTER> to the right of the previous block

Output only the action description, nothing else.
\par\medskip
\textbf{Dynamics Model}
You are a robot dynamics model. Given the current state and an action, predict the next state after executing the action.

Input:
- Initial observation image (starting state)
- Current observation image
- Action to execute

Output:
- The observation image after executing the action

The action space includes picking up letter blocks and placing them on the table to spell words.
\par\medskip
\textbf{Inverse Dynamics Model}
You are a robot inverse dynamics model. Given two consecutive observations and the current subgoal, infer what action was taken between them.

Input:
- Current subgoal description
- Current observation image
- Next observation image

Output:
- The action that was executed to transition from current to next state

Action space includes:
- pick up the block with letter: <LETTER>
- place the block on the table
- place the current block to the right of the previous block
- place the block with letter <LETTER> on the table
- place the block with letter <LETTER> to the right of the previous block

Output only the action description, nothing else.

\par\medskip
\textbf{Optimal Value Model}
You are a robot progress evaluator. Given previous and current observations along with a goal, evaluate the progress towards achieving the goal.

For level 2 tasks:
- Goal is a specific action (e.g., "pick up the block with letter: <A>")

For level 1 tasks:
- Goal is a composite action (e.g., "place the block with letter <A> to the right of the previous block")

For level 0 tasks:
- Goal is the overall task (e.g., "spell the word: CAT using the blocks on the table")

Output:
- 0: No progress (current state is similar to previous state)
- 1: Progress (moved closer to the goal but not yet achieved)
- 2: Achieved (goal has been accomplished)

Output only the number (0, 1, or 2), nothing else.
\end{tcolorbox}

\section{Additional Experiments}

\subsection{Stage-wise Scores of Real-World Tasks}
To provide a granular understanding of policy behavior beyond binary success rates and holistic scores, 
we report the Stage-wise Score for all real-world experiments. 
This metric, denoted as the $i^{th}$ Stage Score, represents the probability of successfully completing the $i^{th}$ subgoal within a long-horizon task.

\textbf{Main Results.} As shown in Table~\ref{tab:results on stage score}, baselines exhibit a sharp temporal decay in performance, especially in \textit{unseen} settings.
In the \textit{Unseen Rearrange Objects} task, while baselines demonstrate competitive capability in the initial phase, their performance collapses rapidly by Stage 3.
In contrast, Anticipation VLA sustains robust completion rates throughout the trajectory (retaining 0.41 at Stage 3), confirming that our recursive anticipation mechanism effectively mitigates compounding errors in long-horizon manipulation.

\begin{table}[htbp]
\centering
{%
\begin{tabular}{llcccccc}
\toprule
\multirow{2}{*}{\textbf{Setting}} & \multirow{2}{*}{\textbf{Model}} & \multicolumn{5}{c}{\textbf{$i^{th}$ Stage Score}} & \multirow{2}{*}{\textbf{SR.} $\uparrow$} \\
\cmidrule(lr){3-7}
 & & 1 & 2 & 3 & 4 & 5 & \\
\midrule
\multicolumn{8}{c}{\textit{\textbf{Task: Rearrange Objects}}} \\
\midrule
\multirow{3}{*}{Seen} 
 & $\pi_{0.5}$ & \textbf{0.95} & \textbf{0.84} & 0.29 & 0.23 & 0.00 & 0.30 \\
 & $\pi_{0.5}$+VLM & 0.80 & 0.63 & 0.29 & 0.15 & 0.00 & 0.20 \\
 & Anticipation-VLA & \textbf{0.95} & 0.79 & \textbf{0.59} & \textbf{0.46} & \textbf{0.43} & \textbf{0.50} \\
\cmidrule{1-8} 

\multirow{3}{*}{Unseen} 
 & $\pi_{0.5}$ & \textbf{0.90} & 0.47 & 0.12 & 0.00 & 0.00 & 0.15 \\
 & $\pi_{0.5}$+VLM & 0.80 & 0.53 & 0.12 & 0.08 & 0.00 & 0.15 \\
 & Anticipation-VLA & \textbf{0.90} & \textbf{0.74} & \textbf{0.41} & \textbf{0.31} & \textbf{0.29} & \textbf{0.40} \\

\midrule
\multicolumn{8}{c}{\textit{\textbf{Task: Spell Words}}} \\
\midrule
\multirow{3}{*}{Seen} 
 & $\pi_{0.5}$ & 0.60 & 0.20 & 0.05 & 0.11 & 0.00 & 0.00 \\
 & $\pi_{0.5}$+VLM & 0.65 & 0.30 & 0.10 & 0.00 & 0.00 & 0.05 \\
 & Anticipation-VLA & \textbf{0.95} & \textbf{0.75} & \textbf{0.55} & \textbf{0.44} & \textbf{0.14} & \textbf{0.40} \\
\cmidrule{1-8}

\multirow{3}{*}{Unseen} 
 & $\pi_{0.5}$ & 0.35 & 0.00 & 0.00 & 0.00 & 0.00 & 0.00 \\
 & $\pi_{0.5}$+VLM & 0.45 & 0.10 & 0.00 & 0.00 & 0.00 & 0.00 \\
 & Anticipation-VLA & \textbf{0.90} & \textbf{0.55} & \textbf{0.35} & \textbf{0.10} & \textbf{0.14} & \textbf{0.30} \\
\bottomrule
\end{tabular}
}
\caption{Stage-wise performance comparison on Real-World Tasks.}
\label{tab:results on stage score}
\end{table}

\textbf{Ablation Analysis.}  
The breakdown in Table~\ref{tab:ablation on stage score} further clarifies the failure modes of each ablation variant. Notably, the \textit{w/o recursive} variant shows a distinct performance drop-off in the later stages, validating the hypothesis that fixed-level generation lacks the adaptability required for deep subgoal planning. Complementing this, multimodal subgoals provide essential grounding: ablating visual or textual predictions precipitates a rapid collapse, stemming from the deprivation of visual guidance from future goals or textual instructions, respectively.

\begin{table}[htbp]
\centering
{%
\begin{tabular}{llcccccc}
\toprule
\multirow{2}{*}{\textbf{Setting}} & \multirow{2}{*}{\textbf{Model}} & \multicolumn{5}{c}{\textbf{$i^{th}$ Stage Score}} & \multirow{2}{*}{\textbf{SR.} $\uparrow$} \\
\cmidrule(lr){3-7}
 & & 1 & 2 & 3 & 4 & 5 & \\
\midrule
\multicolumn{8}{c}{\textit{\textbf{Task: Rearrange Objects}}} \\
\midrule
\multirow{3}{*}{Seen} 
 & Anticipation-VLA & \textbf{0.95} & \textbf{0.79} & \textbf{0.59} & \textbf{0.46} & \textbf{0.43} & \textbf{0.50} \\
& w/o subgoal image & 0.85 & 0.53 & 0.41 & 0.23 & 0.00 & 0.25 \\
 & w/o subgoal text & 0.80 & 0.74 & 0.53 & 0.31 & 0.00 & 0.35 \\
 & w/o recursive & \textbf{0.95} & 0.68 & 0.24 & 0.23 & 0.14 & 0.20 \\
\cmidrule{1-8} 

\multirow{3}{*}{Unseen} 
 & Anticipation-VLA & 0.90 & \textbf{0.74} & \textbf{0.41} & \textbf{0.31} & \textbf{0.29} & \textbf{0.40}\\
  & w/o subgoal image & 0.70 & 0.47 & \textbf{0.41} & 0.08 & 0.14 & 0.25 \\
 & w/o subgoal text & \textbf{0.95} & 0.47 & 0.29 & 0.23 & 0.14 & 0.25 \\
 & w/o recursive & 0.85 & 0.47 & 0.18 & 0.08 & 0.00 & 0.20 \\
\midrule
\multicolumn{8}{c}{\textit{\textbf{Task: Spell Words}}} \\
\midrule
\multirow{3}{*}{Seen} 
 & Anticipation-VLA & \textbf{0.95} & \textbf{0.75} & \textbf{0.55} & \textbf{0.44} & \textbf{0.14} & \textbf{0.40} \\
  & w/o subgoal image & \textbf{0.95} & 0.40 & 0.10 & 0.00 & 0.00 & 0.10 \\
 & w/o subgoal text & 0.90 & 0.60 & 0.30 & 0.11 & 0.00 & 0.20 \\
 & w/o recursive & 0.80 & 0.45 & 0.15 & 0.00 & 0.00 & 0.10 \\
\cmidrule{1-8}

\multirow{3}{*}{Unseen} 
 & Anticipation-VLA & \textbf{0.90} & \textbf{0.55} & \textbf{0.35} & \textbf{0.10} & \textbf{0.14} & \textbf{0.30} \\
  & w/o subgoal image & 0.50 & 0.15 & 0.00 & 0.00 & 0.00 & 0.00 \\
 & w/o subgoal text & 0.65 & 0.50 & 0.05 & \textbf{0.10} & 0.00 & 0.05 \\
 & w/o recursive & 0.55 & 0.15 & 0.10 & 0.00 & 0.00 & 0.05 \\
\bottomrule
\end{tabular}
}
\caption{Ablation analysis on Real-World Tasks using Stage-wise Scores.}
\label{tab:ablation on stage score}
\end{table}

\textbf{Generalization Analysis.} 
Table~\ref{tab:generalization on stage score} highlights the robustness of our model under severe distribution shifts. 
In the challenging task \textit{Spell Words}, baselines scarcely manage to complete a single stage and fail completely by Stage 2.
Conversely, Anticipation VLA maintains a strong start and continues to make meaningful progress through intermediate stages.

\begin{table}[htbp]
\centering
{%
\begin{tabular}{llcccccc}
\toprule
\multirow{2}{*}{\textbf{Setting}} & \multirow{2}{*}{\textbf{Model}} & \multicolumn{5}{c}{\textbf{$i^{th}$ Stage Score}} & \multirow{2}{*}{\textbf{SR.} $\uparrow$} \\
\cmidrule(lr){3-7}
 & & 1 & 2 & 3 & 4 & 5 & \\
\midrule
\multicolumn{8}{c}{\textit{\textbf{Task: Rearrange Objects}}} \\
\midrule
\multirow{3}{*}{Object} 
 & $\pi_{0.5}$ & 0.70 & 0.40 & 0.00 & 0.00 & 0.00 & 0.10 \\
 & $\pi_{0.5}$+VLM & 0.70 & 0.40 & 0.20 & 0.00 & 0.00 & 0.00 \\
 & Anticipation-VLA & \textbf{0.90} & \textbf{0.70} & \textbf{0.30} & \textbf{0.17} & \textbf{0.50} & \textbf{0.30} \\
\cmidrule{1-8} 

\multirow{3}{*}{Background} 
 & $\pi_{0.5}$ & 0.70 & 0.40 & 0.20 & 0.00 & 0.00 & 0.10 \\
 & $\pi_{0.5}$+VLM & 0.80 & 0.60 & 0.10 & 0.00 & 0.00 & 0.10 \\
 & Anticipation-VLA & \textbf{0.90} & \textbf{0.70} & \textbf{0.50} & \textbf{0.33} & 0.00 & \textbf{0.20} \\
\midrule
\multicolumn{8}{c}{\textit{\textbf{Task: Spell Words}}} \\
\midrule
\multirow{3}{*}{Object} 
 & $\pi_{0.5}$ & 0.30 & 0.00 & 0.00 & 0.00 & 0.00 & 0.00 \\
 & $\pi_{0.5}$+VLM & 0.30 & 0.20 & 0.00 & 0.00 & 0.00 & 0.00 \\
 & Anticipation-VLA & \textbf{0.80} & \textbf{0.40} & \textbf{0.20} & 0.00 & 0.00 & \textbf{0.10} \\
\cmidrule{1-8} 

\multirow{3}{*}{Background} 
 & $\pi_{0.5}$ & 0.30 & 0.10 & 0.00 & 0.00 & 0.00 & 0.00 \\
 & $\pi_{0.5}$+VLM & 0.40 & 0.00 & 0.00 & 0.00 & 0.00 & 0.00 \\
 & Anticipation-VLA & \textbf{0.80} & \textbf{0.40} & \textbf{0.20} & \textbf{0.17} & 0.00 & \textbf{0.20} \\
\bottomrule
\end{tabular}
}
\caption{Generalization results on Real-World Tasks using Stage-wise Scores.}
\label{tab:generalization on stage score}
\end{table}

\subsection{Qualitative Results of Subgoal Stack}
\label{app:goal_stack_vis}
\begin{figure}[htbp]
    \centering
    \includegraphics[width=0.8\linewidth]{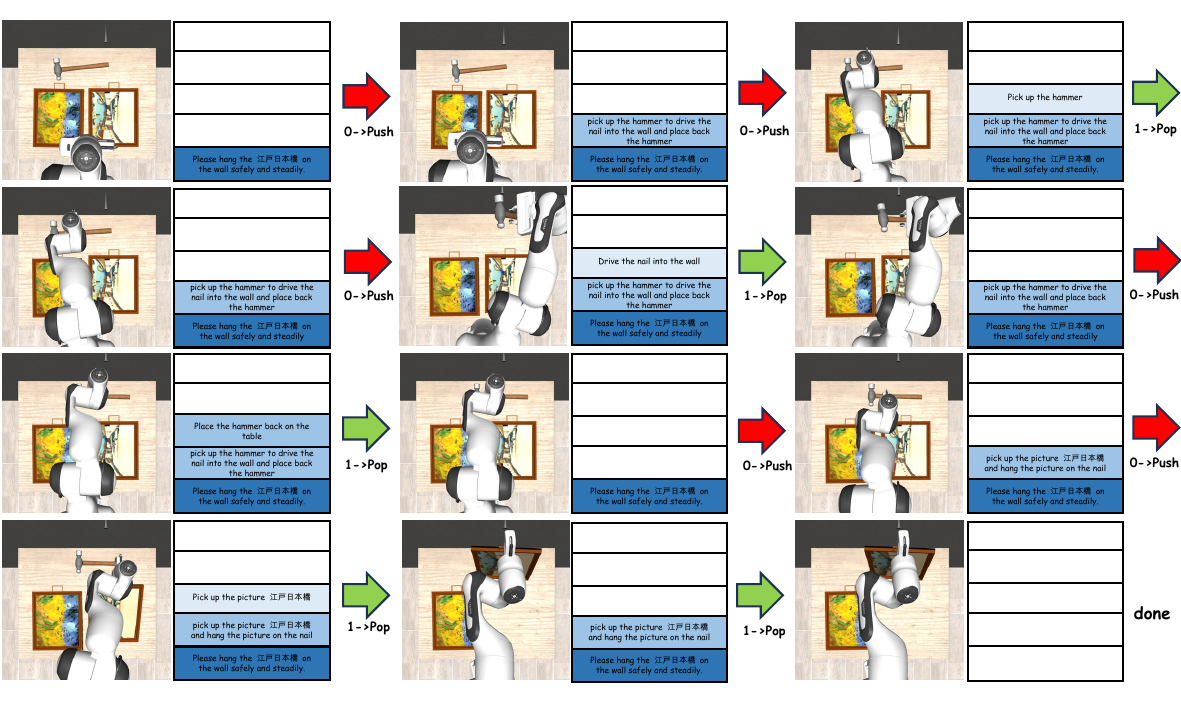}
    \caption{Execution process of the first simulated task in the simulator VLAbench. Given a high-level instruction, the agent sequentially completes sub-tasks such as using the hammer, driving the nail, placing the tool back, and hanging the picture, illustrating the full task workflow under long-horizon execution.}
    \label{fig:libero_s}
\end{figure}

\begin{figure}[htbp]
    \centering
    \includegraphics[width=0.8\linewidth]{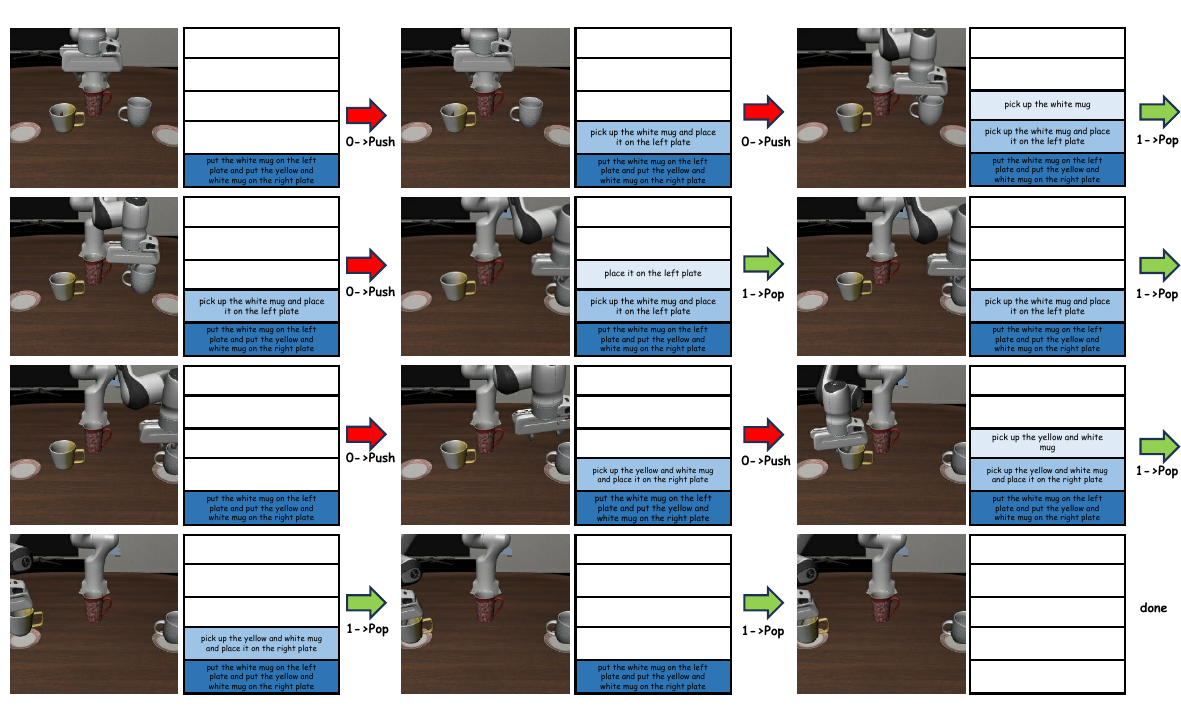}
    \caption{Execution process of the second simulated task in the simulator LIBERO. The agent follows the instruction to rearrange objects by sequentially picking up the white mug and placing it on the left plate, followed by moving the yellow-and-white mug to the right plate, demonstrating the complete task workflow in the simulator.}
    \label{fig:vlabench_s}
\end{figure}

\begin{figure}[htbp]
    \centering
    \includegraphics[width=0.8\linewidth]{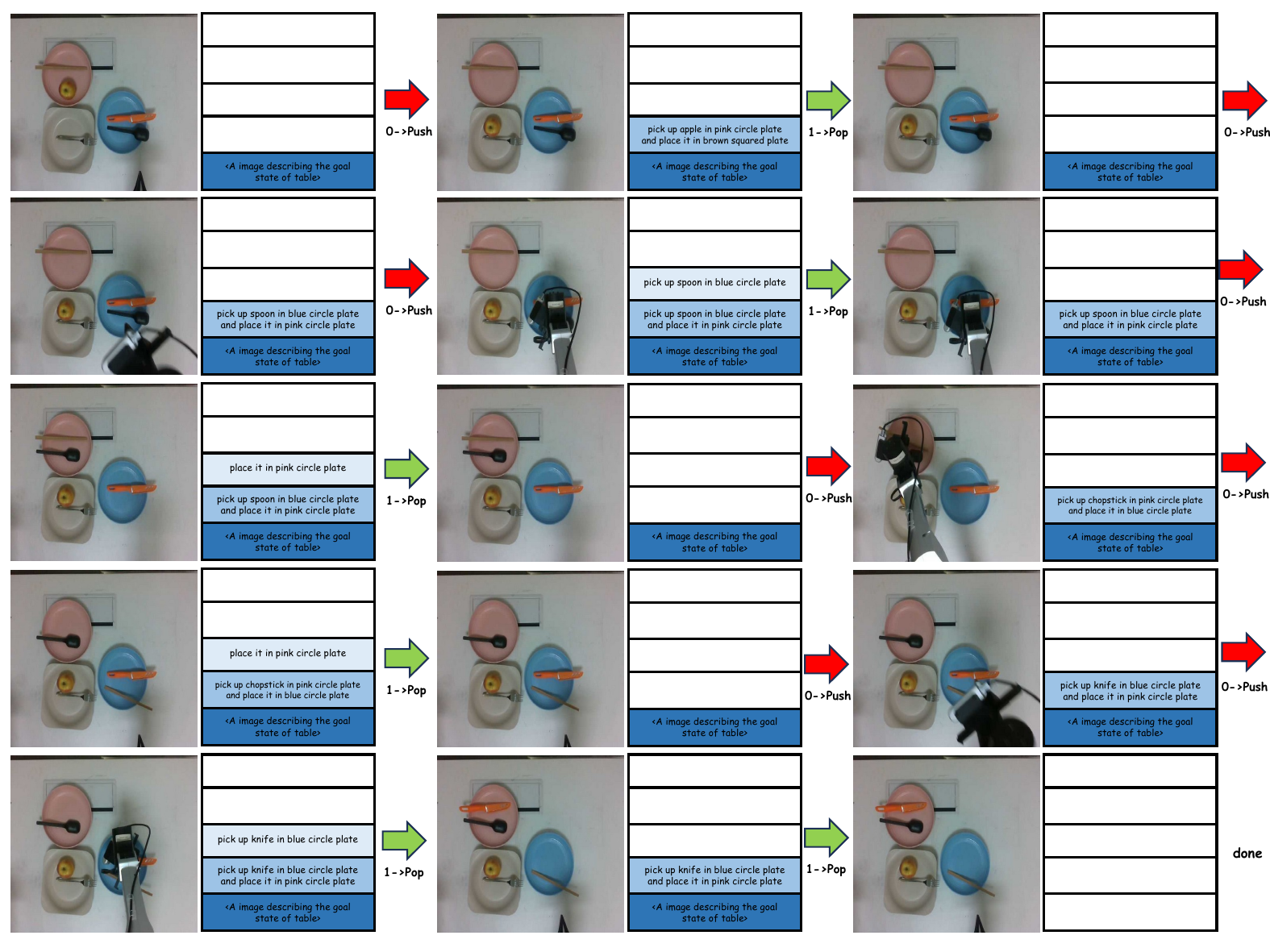}
    \caption{Execution process of the first real-world task. Given a goal state specified by a target image, the robot sequentially rearranges multiple objects by picking and placing items such as the apple, spoon, chopstick, and knife into their designated plates, illustrating the complete task workflow in the real-world setting.}
    \label{fig:pick_s}
\end{figure}

\begin{figure}[htbp]
    \centering
    \includegraphics[width=0.8\linewidth]{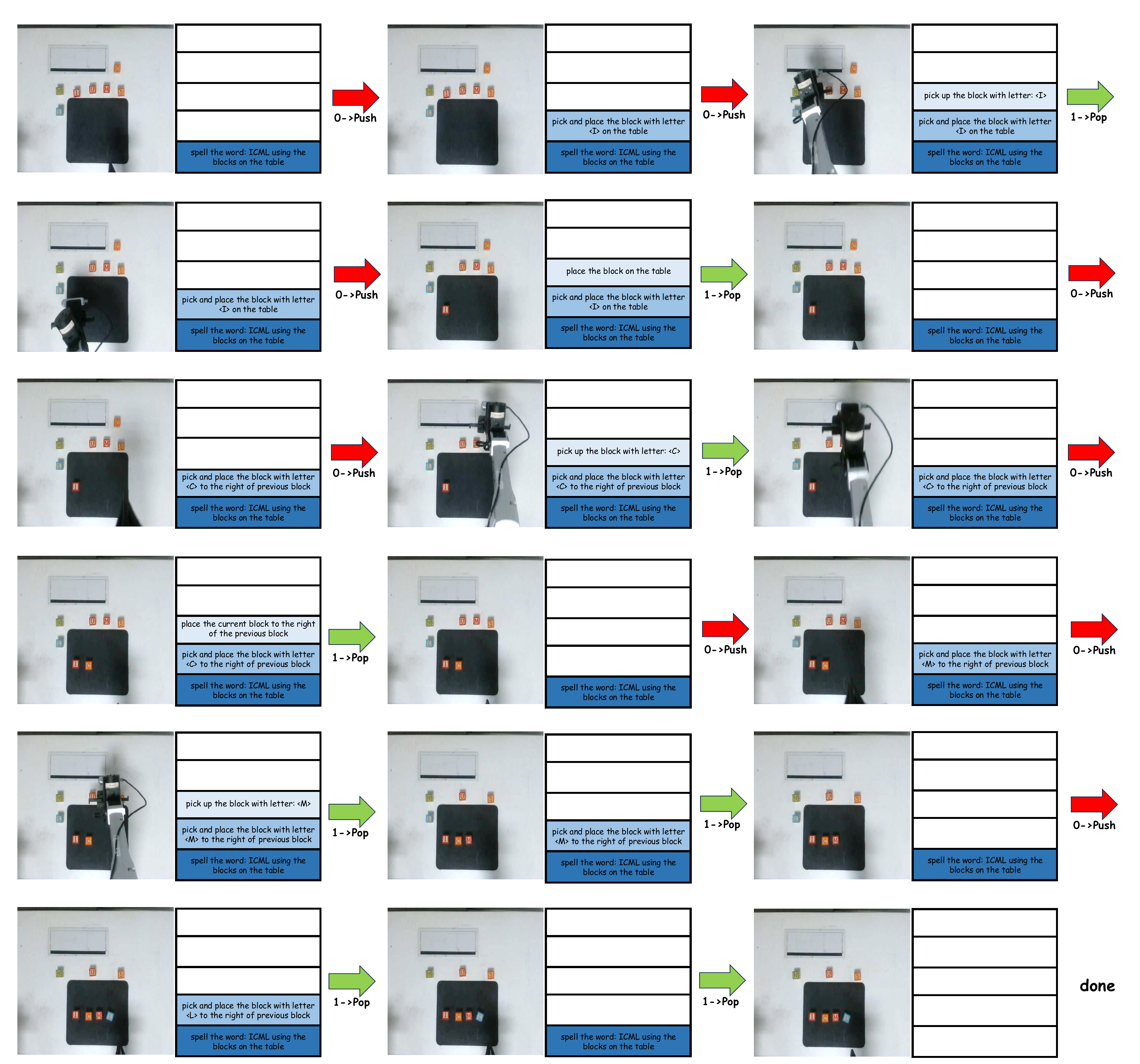}
    \caption{Execution process of the second real-world task. Given a language instruction to spell a target word, the robot sequentially picks and places letter blocks in the correct order (e.g., “I–C–M–L”), demonstrating the complete task workflow for long-horizon execution in the real-world setting.}
    \label{fig:word_S}
\end{figure}

We present, in Figures ~\ref{fig:libero_s}, ~\ref{fig:vlabench_s}, ~\ref{fig:pick_s}, and ~\ref{fig:word_S}, a possible stack evolution of Anticipation across four benchmarks—VLABench, LIBERO,Rearrange Objects, and Spell Words respectively. For each subfigure, the left shows the current observation, and the right shows the corresponding stack state, with the stack top as the current subgoal received by the low-level VLA. Note that we omit cases where the output of Sufficient Progress will make the stack remain unchanged; We only consider two cases: the Insufficient Progress pushing a new subgoal (marked in red, with 0)  and the Goal Achievement popping completed subgoals (marked in green, with 1).
